\documentclass[10pt,journal,compsoc]{IEEEtran}

\ifCLASSOPTIONcompsoc
  \usepackage[nocompress]{cite}
\else
  \usepackage{cite}
\fi
\ifCLASSINFOpdf
\else
\fi

\usepackage{amsmath}
\usepackage{amssymb}
\usepackage{algorithmic}
\usepackage{algorithm}
\usepackage{enumerate}
\usepackage{graphicx}
\usepackage{color}
\usepackage{tikz}
\usepackage{colortbl}

\ifCLASSOPTIONcompsoc
  \usepackage[caption=false,font=footnotesize,labelfont=sf,textfont=sf]{subfig}
\else
  \usepackage[caption=false,font=footnotesize]{subfig}
\fi

\usepackage{booktabs}
\usepackage{fixltx2e}
\usepackage{dblfloatfix}
\usepackage{url}
\usepackage{bm}
\usepackage{hyperref}
\usepackage{multirow,tabularx}
\newcolumntype{C}{>{\centering\arraybackslash}X}

\usepackage{ulem}

\usepackage{amsthm}

\theoremstyle{plain}
\newtheorem{prop}{Proposition}
\theoremstyle{definition}

\newtheorem{thm}{Characteristic}

\begin{document}
%
\title{\huge FINER++: Building a Family of Variable-periodic Functions for Activating Implicit Neural Representation }

%
%
%
%

\author{Hao Zhu$^\dagger$, Zhen Liu$^\dagger$, Qi Zhang, Jingde Fu, Weibing Deng, Zhan Ma, Yanwen Guo, Xun Cao$^*$
\IEEEcompsocitemizethanks{\IEEEcompsocthanksitem The first two authors contributed equally. X. Cao is the corresponding author (caoxun@nju.edu.cn)
\IEEEcompsocthanksitem H. Zhu, Z. Liu, Z. Ma, X. Cao are with the School of Electronic Science and Engineering, Nanjing University, Nanjing, 210023, China.
\IEEEcompsocthanksitem Q. Zhang is with the Vivo, Hangzhou, 310013, China.
\IEEEcompsocthanksitem J. Fu, W. Deng are with the School of Mathematics, Nanjing University, Nanjing, 210023, China.
\IEEEcompsocthanksitem Y. Guo is with the Department of Computer Science and Technology, Nanjing University, Nanjing, 210023, China.
}
\thanks{Manuscript received April 19, 2005; revised August 26, 2015.}}

%
%

\markboth{Journal of \LaTeX\ Class Files,~Vol.~14, No.~8, August~2015}%
{Shell \MakeLowercase{\textit{et al.}}: Bare Advanced Demo of IEEEtran.cls for IEEE Computer Society Journals}

\IEEEtitleabstractindextext{%
\begin{abstract}
Implicit Neural Representation (INR), which utilizes a neural network to map coordinate inputs to corresponding attributes, is causing a revolution in the field of signal processing. However, current INR techniques suffer from the ``frequency''-specified spectral bias and capacity-convergence gap, resulting in imperfect performance when representing complex signals with multiple ``frequencies''. We have identified that both of these two characteristics could be handled by increasing the utilization of definition domain in current activation functions, for which we propose the FINER++ framework by extending existing periodic/non-periodic activation functions to variable-periodic ones. By initializing the bias of the neural network with different ranges, sub-functions with various frequencies in the variable-periodic function are selected for activation. Consequently, the supported frequency set can be flexibly tuned, leading to improved performance in signal representation. We demonstrate the generalization and capabilities of FINER++ with different activation function backbones (Sine, Gauss. and Wavelet) and various tasks (2D image fitting, 3D signed distance field representation, 5D neural radiance fields optimization and streamable INR transmission), and we show that it improves existing INRs. \textit{Project page:} \url{https://liuzhen0212.github.io/finerpp/}

\end{abstract}

\begin{IEEEkeywords}
Implicit neural representation, Variable-periodic activation functions, Spectral bias
\end{IEEEkeywords}}

\maketitle

\IEEEdisplaynontitleabstractindextext

%
\IEEEpeerreviewmaketitle

\ifCLASSOPTIONcompsoc
\IEEEraisesectionheading{\section{Introduction}\label{sec:introduction}}
\else
\section{Introduction}
\label{sec:introduction}
\fi

\IEEEPARstart{T}{he} way a signal is represented is the foundation of all the following problems and determines the paradigm for solving them. Traditional representations, such as the image matrices, point cloud or volumes~\cite{tewari2022advances}, focus on recording the elements individually and have offered significant contributions in history. However, this representation is increasingly inadequate for addressing the numerous inverse problems arising in modern times, such as neural rendering~\cite{tewari2022advances}, inverse imaging~\cite{xie2022neural} and simulations~\cite{karniadakis2021physics, deng2023fluid}. On the contrary, the implicit neural representation (INR)~\cite{sitzmann2020implicit}, which characterizes a signal by preserving the mapping function from the coordinates to corresponding attributes using neural networks, is gaining increasing attentions thanks to the advantages of querying attributes continuously and incorporating differentiable physical process seamlessly. As a result, INR has found widespread application in solving domain-specific inverse problems~\cite{karniadakis2021physics,hao2022physics}, particularly in cases where large-scale paired datasets are unavailable, and only measurements and forward physical process are provided.


However, existing INR techniques suffer the well-known spectral-bias~\cite{rahaman2019spectral,yuce2022structured}, \textit{i.e.}, the low-frequency components of the signal are more easily to be learned. To address this bias, the positional encoding~\cite{mildenhall2020nerf,tancik2020fourier} which aims at embedding multiple orthogonal Fourier or Wavelet bases~\cite{fathony2020multiplicative} into the subsequent network is proposed. However, a significant challenge arises from the fact that the frequency distribution of a signal to be inversely solved is often unknown. This can potentially lead to a mismatch between the pre-defined bases' frequency set and the characteristics of the signal itself, resulting in an imperfect representation~\cite{xie2023diner}. Apart from the pre-defined frequency set, there is a growing focus within the research community on automatic frequency tuning, achieved through the use of periodic~\cite{sitzmann2020implicit} or non-periodic activation~\cite{ramasinghe2022beyond,saragadam2023wire} functions. Nevertheless, these specified-designed activation functions merely alleviate the spectral-bias to a ``frequency''-specified one and inappropriate parameters initialization often results in a capacity-convergence gap.



\begin{figure*}[!t]
\begin{center}
\centering
\subfloat[Comparisons of activation functions]{
	\label{fig:funct_curves_cmp}
	\includegraphics[width=0.606\linewidth]{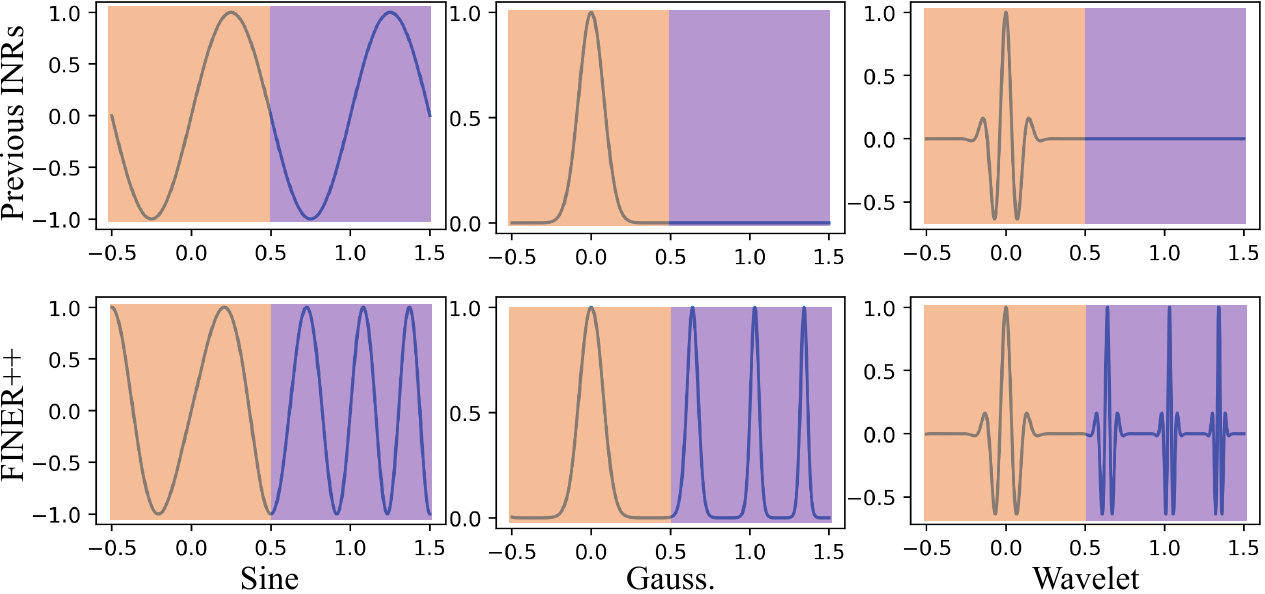}
}
\subfloat[Comparisons of training curves]{
	\label{fig:res_behavior_curves}
	\includegraphics[width=0.394\linewidth]{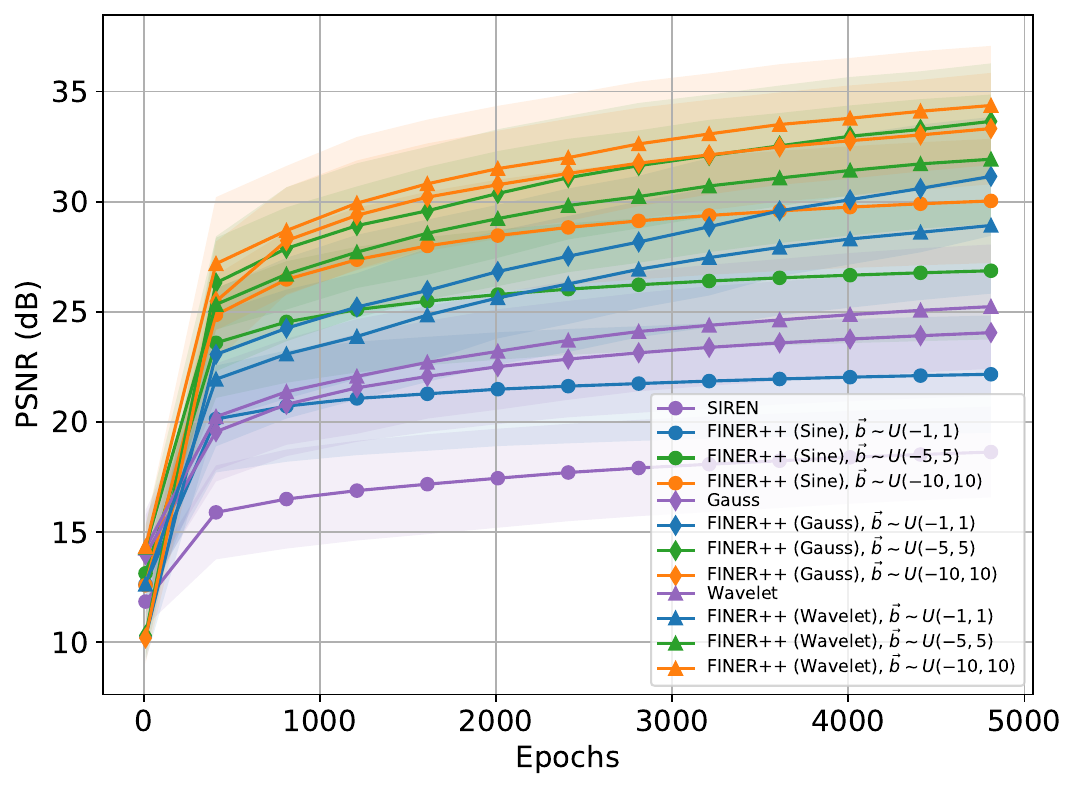}
}
\end{center}
\caption{FINER++ framework for INR. We observe that the supported frequency set in classical INRs is limited due to the under-utilization of activation functions' definition domain, \textit{i.e.}, they mainly employ the central region near the origin point. To overcome this limitation, we propose the FINER++ framework by extending the activation functions from periodic/non-periodic functions to variable-periodic ones. This innovation allows for tuning the supported frequency set by adjusting the initialization range of the bias vector in the neural network. (a) visualizes the selected narrow activation functions in classical activation Sine, Gauss. and Wavelet alongside our proposed variable-periodic ones with different bias settings (purple areas). (b) plots the training curves of previous INRs and FINER++, demonstrating the impact of different initializations applied to the bias vector $\vec{b}$ (see Sec.~\ref{sec:img_diff_bias} for more details). 
}
\label{fig:firstimg}
\end{figure*}

These problems are closely related to a same phenomenon, \textit{i.e.}, the under-utilization of definition domain in the used activation functions. While the widespread used activation functions have an infinite domain, the non-linear components are often concentrated on a small region centered around the origin and the input values are mostly dropped into these areas in practical applications. By ``interspersing'' narrow activation functions with different frequencies along the full definition domain and then selecting the ideal one by controlling the range of input values, the supported frequency set could be significantly expanded, resulting in improved expressive power of current INRs. Following this idea, we propose the FINER++, which is a universal framework to extend current activation functions to their corresponding variable-periodic versions, where the local non-linear components are repeated many times along the $x$-axis and the scale parameters of each local components are changed.

Different from previous explorations~\cite{sitzmann2020implicit,ramasinghe2022beyond,saragadam2023wire} which focus on optimizing the weight matrix for manipulating frequency candidates with better matching degree, FINER++ opens a novel way to achieve frequency tuning by modulating the bias vector, or in other words, the phase of the variable-periodic activation functions. 
We demonstrate that no matter which backbone activation function is used, both the shift-invariance and eigenvalues distribution of neural tangent kernel (NTK) in its FINER++'s version can be enhanced (see Figs~\ref{fig:funct_curves_cmp}, \ref{fig:NTK_vis}) by increasing the initialization range of bias vector, thus the spectral bias could be flexibly tuned and the capacity-convergence gap could be significantly alleviated.
To verify the performance, extensive experiments are conducted on 2D image fitting, 3D signed distance field representation, 5D neural radiance field optimization and streamable implicit neural representation.

This work extends our preliminary exploration as presented in CVPR’24~\cite{liu2024finer}. In comparison with the conference version, we first derive and summarize three key characteristics of INRs, \textit{i.e.}, under-utilized definition domain in activation function, ``frequency''-specified spectral bias and capacity-convergence gap. Subsequently, we introduce a universal framework that extends existing periodic and non-periodic activation functions to their variable-periodic versions, moving beyond the Sine function. This enhancement improves the utilization rate of the definition domain and thus handling the latter two characteristics. To validate this framework, we re-conducted all experiments from the conference version using a wider range of backbone activation functions, achieving superior performance. Moreover, we demonstrate the enhanced capabilities of FINER++ in a novel task: streamable INR transmission.

The main contributions of the work include,
\begin{enumerate}
    \item We derive and summarize three characteristics of existing INRs, which motivates the design of a family of variable-periodic activation functions.
    \item We build a universal framework FINER++ to improve the performance of current INRs by extending periodic and non-periodic activation functions to their variable-periodic versions.
    \item We propose a novel initialization scheme for FINER++ and prove its effectiveness and efficiency from both the perspectives of geometry and neural tangent kernel.
    \item We substantiate that FINER++ surpasses prior INRs activated with other functions for 2D image fitting, 3D signed distance field representation, 5D neural radiance field optimization and streamable INR transmission.
\end{enumerate}


\section{Related Work}
\subsection{Implicit Neural Representation}
INRs~\cite{sitzmann2020implicit, tancik2020fourier} serve as the foundational building blocks for neural scene representations. These representations are designed to learn continuous functions using a multi-layer perception (MLP) that maps coordinates to visual signals, such as images~\cite{dupont2021coin, lindell2021bacon, xie2023diner}, videos~\cite{kasten2021layered, chen2021nerv}, and 3D scenes~\cite{mildenhall2020nerf, wang2021neus}. Neural Radiance Fields or NeRF \cite{mildenhall2020nerf}, a notable breakthrough in this domain, learns a 5D INR to reconstruct the radiance fields (density and view-dependent color) of a scene. With the widespread application of NeRF and its variants \cite{muller2022instant, barron2021mip} on realistic view synthesis, INRs have rapidly expanded into various fields of vision and signal processing, such as cross-model media representation/compression \cite{gao2022objectfolder, strumpler2022implicit}, neural camera representations \cite{huang2022hdr, huang2023inverting}, microscopy imaging \cite{zhu2022dnf} and partial differential equations solver~\cite{raissi2019physics, karniadakis2021physics}.
Despite the interest and success of these implicit representations, current approaches often suffer from the well-known spectral-bias problem. As a result, the INRs may struggle to capture high-fidelity details of complex signals, leading to suboptimal performance in fitting functions and ineffectiveness in various applications.

\subsection{Spectral-bias Problem}
The issue of spectral-bias in deep learning \cite{rahaman2019spectral} refers to the innate propensity of these models to selectively learn specific patterns or features from input data. In the case of INR-based methods, this problem manifests as a preference for learning low-frequency components of a signal more readily than high-frequency components. To address the spectral-bias problem, several innovative strategies have been proposed for INR-based methods. In particular, spatial encoding is applied to the input data, such as frequency or polynomial decomposition \cite{tancik2020fourier, singh2023polynomial, raghavan2023neural}, high-pass filtering \cite{wu2023neural, fathony2020multiplicative}, to emphasize high-frequency components before feeding into the model. \cite{tancik2020fourier} uses the Fourier features mapped from spatial coordinates as the input of MLPs to improve the performance of INRs in adequately expressing high-frequency information of signals. Additionally, various architectural modifications have been integrated into INRs, including multi-scale or pyramid representations \cite{lindell2021bacon, saragadam2022miner, zhu2023pyramid}, which can aid in capturing both low-frequency and high-frequency components of a scene. \cite{lindell2021bacon} implements a multi-scale network architecture with a band-limited Fourier spectrum to minimize artifacts during the downsampling or upsampling process. However, the frequency distribution of a signal requiring inverse solving is often unknown, making it difficult to design an appropriate representation or model for the signal without prior knowledge of its frequency content.

In addition to the positional encoding with pre-defined frequency bases, there has been a growing interest in the research community for automatic frequency tuning through the use of nonlinearity activation functions \cite{sitzmann2020implicit, chng2022gaussian, ramasinghe2022beyond, saragadam2023wire}. Sitzmann et al.~\cite{sitzmann2020implicit} propose the Sinusoidal Representation Network (SIREN), a method designed to represent complex signals and functions using periodic activation functions, especially sine functions. SIREN has demonstrated its effectiveness in representing intricate details and high-frequency components when compared to traditional activation functions like ReLU or sigmoid. Ramasinghe et al.~\cite{ramasinghe2022beyond} propose the Gaussian-based activation function to improve SIREN's performance with random initializations. Saragadam et al.~\cite{saragadam2023wire} draw inspiration from signal processing literature and propose the Wavelet-based activation functions to improve INR's performance for visual signals. However, INRs activated with different functions could be viewed as different function expansion process such as the Fourier expansion~\cite{benbarka2022seeing}, Gaussian mixture models or Wavelet expansion, and the first layer of these INRs could be viewed as bases encoding layers~\cite{yuce2022structured}, as a result, the problem of requiring prior frequency knowledge also exists.

Apart from INRs with specified-designed encoding layers or activation functions, the spectral bias could also be alleviated by introducing hash-tables between the input coordinates and the neural networks. Muller et al.~\cite{muller2022instant} propose the learnable multi-resolution hash-tables, where each input coordinate are used to index hash-keys which are fed into the subsequent network, then these hash-keys and network parameters are jointly optimized during the training process. Kang et al.~\cite{kang2023pixel} propose the multi-shift windows hash-tables and achieve better performance for solving partial differential equations. Abou et al.~\cite{abou2024particlenerf} extend the multi-resolution hash-tables to dynamic scenes where the default uniform hash grid is changed according to the motions. Zhu et al.\cite{zhu2024disorder} visualize the learned hash keys and highlight the role of the hash table in reorganizing complex signals into simpler low-frequency ones, thereby significantly enhancing the performance of hash-table-based INRs using a compact network. Their subsequent work~\cite{zhu2023rhino} regularizes the noisy interpolation in hash-table-based INR by introducing a continuous analytical function between the input and the network. Although the hash-based INRs outperform function-expansion-based ones, the significantly increased storage overhead for saving hash-tables limits its applications in storage-sensitive tasks such as the compression and transmission.

\section{INR and its Characteristics}
In this section, we first give the definition of INR. Then three characteristics of INR are analysed and explained in detail, \textit{i.e.}, the under-utilized definition domain in activation functions, ``frequency''-specified spectral bias and the capacity-convergence gap.
\label{sec:INR_properties}

\subsection{Pipeline of INRs} 
Given a signal sequence $\{\vec{x}_i,\vec{y}_i\}_{i=1}^{N}$, where $\vec{x}_i$ and $\vec{y}_i$ respectively represent the coordinate and the corresponding attributes of the $i$-th element, $N$ is the number of elements in the signal. INR focuses on pursuing a neural network $f(\vec{x};\theta)$ to characterize the attributes as accurate as possible. Mostly, the multi-layer perceptron (MLP) network is used to model the $f(\vec{x};\theta)$ which could be formulated as follows:
\begin{equation}
    \begin{split}
        \displaystyle
        {\vec{z}}^{\:0} &= \vec{x} \\
        \vec{z}^{\:l} &= \sigma_{p}(W^{l} \vec{z}^{\:l-1} + \vec{b}^{\:l}),\ l=1,2,...,L-1, \\
        f(\vec{x} ; \theta) &= W^{L} \vec{z}^{\:L-1} + \vec{b}^{\:L}
    \end{split}
    \label{eqn:periodic_activation}
\end{equation}
where $\displaystyle \vec{z}^{\:l}$ denotes the output of layer $l$, $\theta=\{W^{l}, \vec{b}^{\:l}\ |\ l=1,2,...,L\}$ are the network parameters to be optimized, $L$ is the number of layers. $\sigma_{p}(x)$ is the activation function, such as the periodic function $\sin(\omega_{0}x)$~\cite{sitzmann2020implicit}, the non-periodic Gaussian function $e^{-(s_{0}x)^{2}}$~\cite{ramasinghe2022beyond} and the Wavalet function $e^{j\omega_0x}e^{-|s_0x|^2}$~\cite{saragadam2023wire}, $p\in\{\omega_0,s_{0},(\omega_0,s_{0}),...\}$ refers to the empirical parameter for controlling the scale of these functions\footnote{In this paper, we indiscriminately use the terms Sine/Gauss./Wavelet to represent Sine/Gauss./Wavelet activation functions and the corresponding INRs.}.

\subsection{Under-utilized Definition Domain in Activation functions}
\label{sec:subsec:under_utilization}
In classical deep learning theory~\cite{goodfellow2016deep}, the input of neural network is often normalized to $[-1,1]$ or other ranges and the network parameters are also initialized in a fixed range following various strategies (\textit{e.g.}, Xavier initialization~\cite{glorot2010understanding} and Kaiming initialization~\cite{he2015delving}). Because of these fixed ranges, the value range of each neuron before activation is often distributed in a small region centered around the origin point. As a result, existing works always focus on designing activation functions with non-linearity at a small region centered around the origin point and ignore a wide range of areas which are far away from the origin point. For example, most of the non-linearity in both Gaussian~\cite{ramasinghe2022beyond} and Wavelet~\cite{saragadam2023wire} functions gathers at the $[-3\frac{1}{\sqrt{s_0}},3\frac{1}{\sqrt{s_0}}]$ according to the Statistics theory~\cite{schervish2012theory}, where $\frac{1}{\sqrt{s_0}}$ refers to the standard deviation of the Gaussian function or the Gaussian component of the Wavelet function. Note that although the non-linearity appears every periodic in $\sin$ function, only non-linearity in the central periodic (\textit{i.e.}, $[-\pi,\pi]$) is used considering the convergence and performance of SIREN~\cite{sitzmann2020implicit}. In summary,

\begin{thm}
\label{thm:under_utilization}
    \textit{Current designs of INRs' activation functions focus solely on non-linearities centered around the origin and the definition domains of these functions are under-utilized.}
\end{thm}


\subsection{``Frequency''-specified Spectral Bias} 
Although the universal approximation theorem has proved the capability of MLP with infinite width and depth for representing any functions, only a series of functions could be represented especially considering the specifically designed activation functions mentioned above. To derive the properties of this series of functions, or in other words, the expressive power of INR\footnote{Note that the expressive power of INR with Sine-based activation has been analysed by Yuce et al.~\cite{yuce2022structured}, we focus on a universal formula with more activations in this paper.}, we follow the idea of viewing INRs as a function expansion process~\cite{benbarka2022seeing}, where the first layer including the activation~\cite{yuce2022structured} is modelled as encoding bases with different parameters. By extending the non-linear activation function $\sigma(x)$ as polynomial functions and assuming the signal attribute is a 1-dimensional scalar, the functions which could be represented by the INR follows the form,
\begin{equation}
    \label{eqn:inr_capicity}
    f(\vec{x};\theta) = \sum_{p'\in \mathcal{F}_{p}} c_{p'} \sigma_{p'} (\vec{x}, \phi_{p'}),\:\:\:\: \ c_{p'}, \phi_{p'} \in \mathbb{R} 
\end{equation}
where $\mathcal{F}_{p}$ is the supported ``frequency'' set defined by the empirical scale parameter $p$ and $\phi_{p'}$ represents the phase of a trigonometric function or the mean of a Gaussian function. Note that the connotation of ``frequency'' changes according to different activation functions, \textit{e.g.}, ``frequency'' refers to the frequency and variance when the Sine and Gaussian functions are used, respectively.

For example, when the periodic function $\sin(\omega_0 x)$, Gaussian function $e^{-(s_{0}x)^{2}}$ and Wavelet function $e^{j\omega_0x}e^{-|s_0x|^2}$ are applied, Eqn.~\ref{eqn:inr_capicity} could be written as,
\begin{equation}
\label{eqn:inr_capicity:indetail}
\begin{aligned}
&f(\vec{x};\theta)=\\
&\begin{cases}
\sum\limits_{\:\:\:\:\:\:\:\:\omega'\in \mathcal{F}_{\omega_0}\:\:\:\:\:\:\:} c_{\omega'} \sin (\omega'\vec{x}+\phi_{\omega'}) & \text{Sine}\\
\sum\limits_{\:\:\:\:\:\:\:\:s' \in \mathcal{F}_{s_0}\:\:\:\:\:\:\:\:} 
    c_{s'} e^ {-(s'(\vec{x}-\mu_{s'}))^2} & \text{Gauss.}\\
\sum\limits_{\omega' \in \mathcal{F}_{\omega_0}, s' \in \mathcal{F}_{s_0}} 
    c_{\omega', s'} e^{j(\omega'\vec{x}+\phi_{\omega'})} e^{-(s'(\vec{x}-\mu_{s'}))^2} & \text{Wavelet}
\end{cases}
\end{aligned}
\end{equation}
where $c_{\omega'},\phi_{\omega'},c_{s'},\mu_{s'},c_{\omega',s'}\in \mathbb{R}$ (Please refer the supplemental material for details of derivation). From Eqn.~\ref{eqn:inr_capicity:indetail}, it is observed that MLPs with specifically designed activation functions may not be able to represent arbitrary functions. As a result, the second characteristic of INR could be summarized as,

\begin{thm}
\label{thm:spectral_bias}
\textit{In INRs, there exists a ``frequency''-specified spectral bias, where the type of ``frequency'' is determined by the chosen activation function, and the supported set of ``frequency'' is determined by the empirical scale parameter within the activation function.}
\end{thm}



\subsection{Capacity-Convergence Gap} 
According to the Eqn.~\ref{eqn:inr_capicity:indetail}, the expressive power of INRs is determined by empirical scale parameters $\omega_0$ and $s_0$. Considering the fact that these empirical parameters play the role of scaling the input values, it seems that they could be removed or set as 1 by default since the scaling role can also be played by changing the initialization ranges of the weight and bias matrixes $\{W^{l},\vec{b}^{l}\}_{l=1,...,L}$.

However, previous studies indicate that specialized empirical scale values and specially designed initialization strategies are crucial. Otherwise, the performance may be significantly reduced. For example, the weight matrix $\{W^{l}\}_{l=1,...,L}$ in SIREN~\cite{sitzmann2020implicit} are initialized following a uniform distribution with the range $[-\sqrt{6/n},\sqrt{6/n}]$, where $n$ is the number of inputs for each neuron. Although other papers~\cite{ramasinghe2022beyond,saragadam2023wire} do not require specified designed initialization schemes, it is found that this performance reduction also exists when setting the scale parameters as 1 and changing the initialization ranges of network parameters. 

\begin{figure}[t] 
    \begin{center}
        \includegraphics[width=\linewidth]{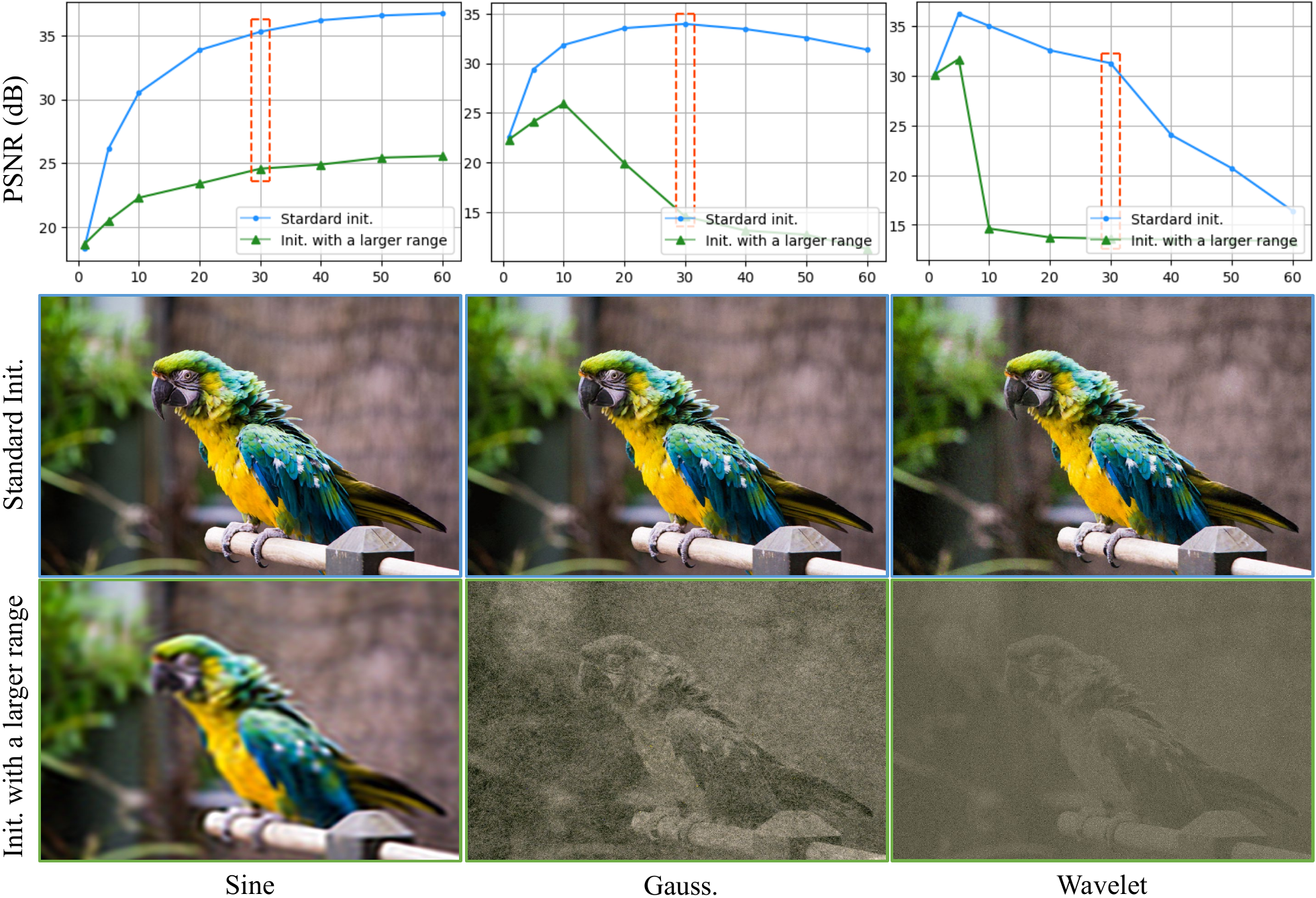}
        \caption{Visualization of capacity-convergence gap in various INRs for fitting a $2K$ image. The performance of various INRs drops significantly when set the empirical scale parameters as 1 and changing the initialization range of network parameters.}
        \label{fig:sec_3.1}
    \end{center}
\end{figure}

Fig.~\ref{fig:sec_3.1} compares the performance of various INRs with different initialization strategies for fitting a high-resolution image. The blue line are the results following the standard form of authors, \textit{i.e.}, default initialization strategies (Gauss. and Wavelet) and $W\sim\mathcal{U}(-\sqrt{6/n},\sqrt{6/n})$ (Sine), then an additional scale parameter is applied to multiply the $W\vec{x}+\vec{b}$. 

On the contrary, the green line refers to the performance by removing the scale parameters in Eqn.~\ref{eqn:periodic_activation} and initializing $\{W,\vec{b}\}$ with a larger range (\textit{i.e.}, the default range multiples the corresponding scale parameter in Gauss. and Wavelet and $W\sim\mathcal{U}(-p\sqrt{6/n},p\sqrt{6/n})$ in Sine). Although they are equivalent in the mathematical form, there is a performance gap in practice (the red dotted boxes in Fig.~\ref{fig:sec_3.1}).

As a result, the initialization range of network parameters $\{W^{l},\vec{b}^{l}\}_{l=1,...,L}$ could not be scaled unboundedly considering the convergence. In summary, there is a capacity-convergence gap in existing INRs that,

\begin{thm}
\label{thm:capacity_gap}
    \textit{The function set that INRs could be represented can be enlarged by increasing the initialization range of network parameters, which violates the rule of guaranteeing the convergence, resulting a performance gap between theory and practice.}
\end{thm}


\section{FINER++: Variable-periodic Activation Functions for INRs}
\label{sec:finer++_frame}
To address the characteristics mentioned above, we propose the FINER++ which utilizes variable-periodic functions for activating INRs.
In this section, we will first introduce a unified framework to extend existing periodic/non-periodic functions to their variable-periodic versions, followed by the proposed initialization scheme. Subsequently, we analyse the behaviors of FINER++'s supported ``frequency'' set and training dynamics under different initialization ranges from both geometric and neural tangent kernel perspectives.
\subsection{Variable-periodic Activation Functions}
Although previous methods have attempted to address the second characteristic (\textit{i.e.}, the ``Frequency''-specified spectral bias) by adjusting scale parameters, none have addressed the other two characteristics. We have found that all three of these inherent characteristics or limitations can be mitigated by introducing variable-periodic activation functions. To achieve this goal, we propose the FINER++ framework, which extends any continuous periodic or non-periodic functions to their corresponding variable-periodic forms.
\begin{equation}
\small
\label{eqn:FINER++_framework}
\begin{aligned}
&\sigma_{p}(x)\rightarrow \sigma_{p'}(x'),\:\:\text{where}\\
&\begin{cases}
    \{x',p'\}=\{(|x|+1)x,p\} & \text{Periodic Func.}\\
    \{x',p'\}=\{\sin (\omega_{f}(|x|+1)x), \frac{p}{\omega_f}\} &\text{Non-Periodic Func.}
\end{cases}
\end{aligned}.
\end{equation}
$\omega_f$ is a parameter to control the overlap between different periods (see Sec.~\ref{sec:omega_f_selection} for details). Specifically, when the Sine ($\sin(\omega_0 x)$), Gauss. ($e^{-(s_{0}x)^{2}}$) and Wavelet ($e^{j\omega_0x}e^{-|s_0x|^2}$) functions are used, their corresponding variable-periodic forms $\sigma_{Sine}$, $\sigma_{Gaus}$ and $\sigma_{Wave}$ are
\begin{equation}
\label{eqn:FINER++_framework:detail}
\begin{aligned}
    \sigma_{Sine}&=\sin(\omega_{0}(|x|+1)x)\\
    \sigma_{Gaus}&= e^{-(\frac{s_{0}},{\omega_f}\sin(\omega_{f}(|x|+1)x))^2}\\
    \sigma_{Wave}&= e^{j\frac{\omega_{0}},{\omega_f}\sin(\omega_{f}(|x|+1)x)}e^{-(\frac{s_{0}}{\omega_f}\sin(\omega_{f}(|x|+1)x))^2}.
\end{aligned}    
\end{equation}
Fig.~\ref{fig:funct_curves_cmp} compares activation functions in previous INRs and their corresponding variable-periodic versions. It is observed that the central non-linear components in original functions are repeated multiple times in FINER++. Because the $|x|+1$ increases with the input variable, the ``frequency'' increases along the $x$-axis in the variable-periodic versions. Note that, Eqn.~\ref{eqn:FINER++_framework} focuses on continuous functions and is not available for piecewise functions (\textit{e.g.}, ReLU).


\subsubsection{Handling INR's Characteristics}
To fully utilize the definition domain of the activation functions, a straight-forward method is changing the initialization ranges of the bias vector $\{\vec{b}^{\:l}\}_{l\in [1,L]}$ of the MLP network, which is equivalent to shift different areas of the activation function to the original point. However, it is meaningless in previous activation functions since the areas far away from the original point are all linear components (\textit{e.g.}, Gaussian and Wavelet) or non-linear components but sharing same properties with the central component (\textit{e.g.}, Sine). 

Different from previous activation functions, because the non-linear components exist everywhere along the $x$-axis in the proposed variable-periodic functions (Fig.~\ref{fig:funct_curves_cmp}), shift different areas to the original point becomes meaningful and could address two limitations mentioned in the above sections. For example, since the ``frequency'' varies from sub-function to sub-function in the FINER++, different bias vector $\vec{b}$ leads to different supported ``frequency'' set $\mathcal{F}_{p}$, resulting in different spectral bias. Furthermore, given fixed scale parameters, the selected non-linear components with larger ``frequency'' means enlarging the scale parameters, thus the capacity-convergence gap could also be mitigated. 

\subsubsection{Selection of $\omega_f$}
\label{sec:omega_f_selection}

\begin{figure}[!t] 
    \begin{center}
        \includegraphics[width=\linewidth]{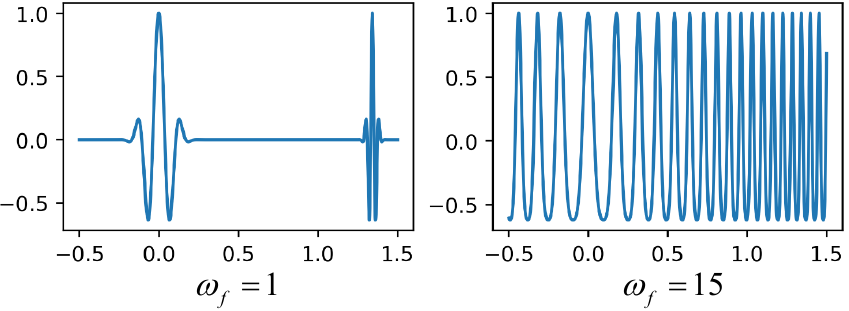}
        \caption{Visualizations of FINER++ (Wavelet) with different $\omega_f$. Inappropriate $\omega_f$ setting results in the problems of ``degeneration'' (left) and ``overlap'' (right).}
        \label{fig:funct_curves_omega_f}
    \end{center}
\end{figure}

As shown in Eqn.~\ref{eqn:FINER++_framework}, the key to extending a function to its variable-periodic version is by composing it with a variable-periodic function, specifically $x'=\sin(\omega_f(|x|+1)x)$. However, due to changes in the operational range of the original functions' non-linear components with different empirical scale parameters parameters p, it is crucial to control the rate of frequency increase parameter $\omega_f$ to avoid issues such as ``degeneration'' and ``overlap''. Fig.~\ref{fig:funct_curves_omega_f} visualizes the curves of FINER++ (Wavelet) with different $\omega_f$ values. It is observed that when $\omega_f$ is small (\textit{e.g.}, $\omega_f=1$ or smaller), FINER++ (Wavelet) gradually degenerates into a standard Wavelet function because the distance between neighboring non-linear components also increases gradually. Conversely, when $\omega_f$ is large (\textit{e.g.}, $\omega_f=15$ or larger), the sidelobes of each non-linear component overlap with neighboring components. We applied Wavelet- and Gaussian-based INRs to represent 2D images and counted the outputs of each neuron across 16 images. It was found that the values before activation mostly fell within an area slightly larger than $[-3\frac{1}{\sqrt{s_0}},3\frac{1}{\sqrt{s_0}}]$ ($\frac{1}{\sqrt{s_0}}$ is the standard derivation of the Gaussian function). As a result, $\omega_f$ was empirically set to 2.5 for all subsequent experiments.

\subsection{Initialization scheme for bias vector} 
\begin{figure} 
    \begin{center}
        \includegraphics[width=\linewidth]{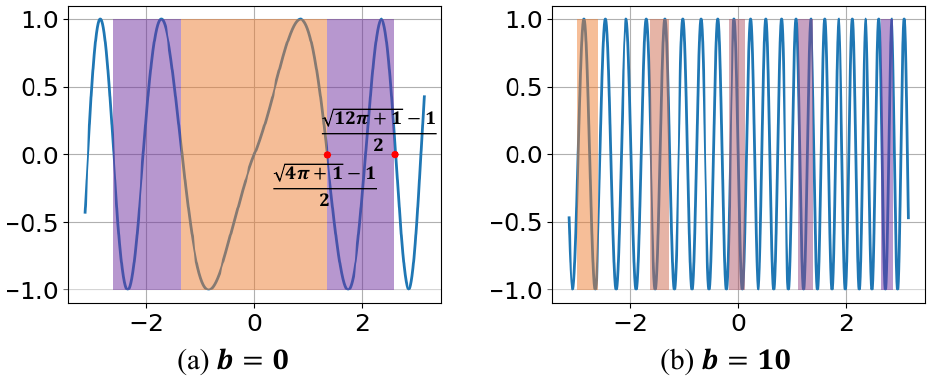}
        \caption{Comparisons of used activation function $\sin((|x|+1)x)$ under different bias $\vec{b}$. More sub-functions with high-frequency are included when $b$ is set with a larger value.}
        \label{fig:sec_3.3}
    \end{center}
\end{figure}

As demonstrated in the previous section, the supported ``frequency'' set of FINER++ can be manipulated by adjusting the ranges of the bias vector $\vec{b}$. However, due to the non-convex nature of variable-periodic activation functions, Eqn.~\ref{eqn:FINER++_framework} exhibits many local minima, and gradient-based optimizations (e.g., SGD or Adam) cannot guarantee moving $\vec{b}$ to the global optimum without proper initialization. Traditional initialization strategies, which typically involve uniform or Gaussian sampling within a narrow region centered around 0, restrict the supported ``frequency'' set of FINER++ to the ``frequency'' of the first non-linear components in the variable-periodic function (\textit{i.e.}, the orange area in Fig.~\ref{fig:sec_3.3}(a)). This limitation results in underutilization of other non-linear components (Fig.~\ref{fig:sec_3.3}(b)), which possess different '‘frequencies'’.

To get rid of the limited supported ``frequency'' set using traditional initialization methods, we derive a novel initialization scheme for $\vec{b}$ for tuning the supported ``frequency'' set flexibly, meanwhile the initialization for $W$ follows the default mechanisms of different INRs~\cite{sitzmann2020implicit,ramasinghe2022beyond,saragadam2023wire}. We propose to initialize $\vec{b}$ following a uniform distribution $\mathcal{U}(-k,k)$ with a larger range $k$ than the default one in traditional methods, 
\begin{equation}
    \vec{b} \sim \mathcal{U}(-k,k), \ k > 0.
\end{equation}
Because the ``frequency'' of different non-linear components increases along the $x$-axis, it is intuitively that the supported ``frequency'' set also increases with a larger $k$ value. To better explain the details of how $k$ influence the behaviors of FINER++, we will focus on the Sine activation function in the following sections.

\begin{figure*}[!t] 
    \begin{center}
        \includegraphics[width=\linewidth]{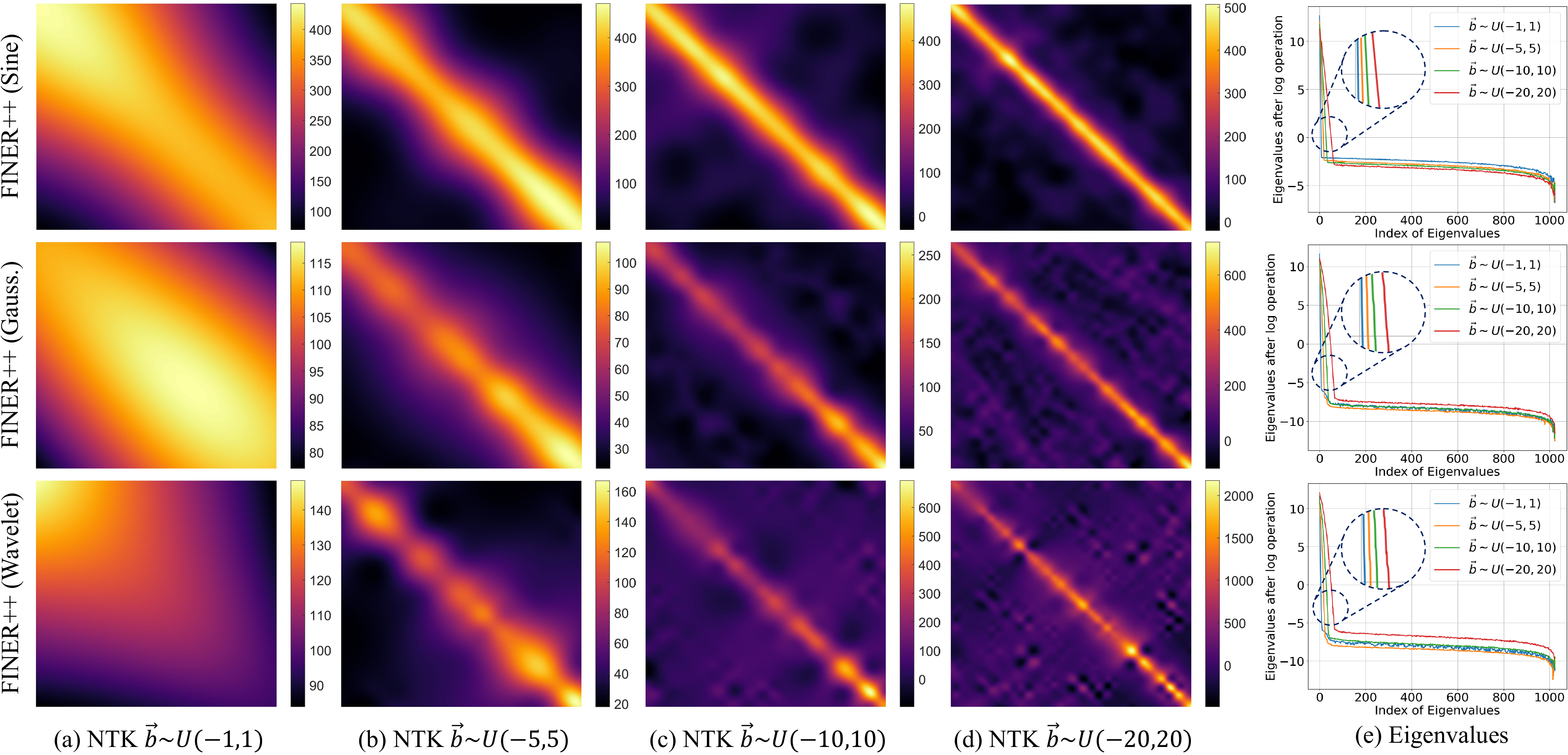}
        \caption{Visualizations of NTKs and the corresponding eigenvalues in FINER++. From top to bottom, the NTKs and NTKs' eigenvalues of FINER++ with sine, Gaussian and Wavelet activation functions are visualized, respectively. From left to right, (a)-(d) visualize the NTKs when $\vec{b}$ is initialized following $\mathcal{U}(-1,1)$, $\mathcal{U}(-5,5)$, $\mathcal{U}(-10,10)$ and $\mathcal{U}(-20,20)$, respectively. (e) plots the corresponding eigenvalues. Because the max eigenvalue is much larger than the smallest one, all eigenvalues are processed by a $\log$ function for visualization.}
        \label{fig:NTK_vis}
    \end{center}
\end{figure*} 

\subsection{Geometrical Perspective} 
Supposing the supported frequency set of SIREN and FINER++ (Sine) are $\mathcal{F}_{\omega_0}$ and $\mathcal{F}_{\omega_0,k}$, respectively. To analyse their relationship, let us start from the simplest case. 

\vspace{0.2cm}
\noindent \textbf{$k$ is close to the origin point.} Because the initialization for $W$ follows \cite{sitzmann2020implicit}, the term $W\vec{x}$ has similar distribution with the one in SIREN, that $|W\vec{x}|\leq \pi$. As a result, the term $|(|W\vec{x}+\vec{b}|+1)(W\vec{x}+\vec{b})|$ drops into the area of $[-\pi^2-\pi,\pi^2+\pi]$. As shown in Fig.~\ref{fig:sec_3.3}(a), the activation function $\sin((|W\vec{x}+\vec{b}|+1)(W\vec{x}+\vec{b}))$ mainly spans two narrow sub-functions with different frequencies. For the points dropped into the first sub-function (\textit{i.e.}, $|W\vec{x}|\leq\frac{\sqrt{4\pi+1}-1}{2}$, the orange areas in Fig.~\ref{fig:sec_3.3}(a)), the supported frequency set $\mathcal{F}_{\omega_0,k}^{1}$ is slightly larger than $\mathcal{F}_{\omega_0}$, and could be approximated as
\begin{equation}
    \mathcal{F}_{\omega_0,k}^{1} \approx \left\{ \frac{2\pi}{\sqrt{4\pi+1}-1} \omega \ | \ \omega\in \mathcal{F}_{\omega_0} \right\}, 
\end{equation}
where the term $\frac{2\pi}{\sqrt{4\pi+1}-1}$ measures the frequency changes from the standard $\sin$ function to the first function in $\sin((|x|+1)x)$. Considering the fact that the range $[-\pi, \pi]$ of $W\vec{x}$ is continuous, \textit{i.e.}, every value could be produced, the $\mathcal{F}_{\omega_0}$ is also a continuous set, as a result,
\begin{equation}
\label{eqn:siren_in_finer}
    \mathcal{F}_{\omega_0} \subset \mathcal{F}_{\omega_0,k}^{1}.
\end{equation}

For the points dropped into the second sub-function (\textit{i.e.}, $\frac{\sqrt{4\pi+1}-1}{2} \leq| W\vec{x}|\leq \frac{\sqrt{12\pi+1}-1}{2}$, the purple areas in Fig.~\ref{fig:sec_3.3}(a)), the supported frequency set $\mathcal{F}_{\omega_0,k}^{2}$ differs from $\mathcal{F}_{\omega_0,k}^{1}$ since the base frequency of the used activation changes. Compared with the $\mathcal{F}_{\omega_0,k}^{1}$ in the first sub-function, the $\mathcal{F}_{\omega_0,k}^{2}$ in the second one could be approximated as,
\begin{equation}
    \mathcal{F}_{\omega_0,k}^{2} \approx \left\{\frac{\sqrt{4\pi+1}-1}{\sqrt{12\pi+1}-\sqrt{4\pi+1}}\omega\ | \ \omega\in\mathcal{F}_{\omega_0,k}^{1}\right\},
\end{equation}
where $\frac{\sqrt{4\pi+1}-1}{\sqrt{12\pi+1}-\sqrt{4\pi+1}}$ characterizes the scale of frequency changes from the first sub-function to the second one. As a result, the supported frequency set $\mathcal{F}_{\omega_0,k}$ for $k$ is close to the origin is,
\begin{equation}
    \mathcal{F}_{\omega_0,k} = \mathcal{F}_{\omega_0,k}^{1} \cup \mathcal{F}_{\omega_0,k}^{2}.
\label{eqn:freq_set_small_b}
\end{equation}
Combining the Eqn.~\ref{eqn:siren_in_finer}, the following equation holds,
\begin{equation}
\label{eqn:siren_in_finer_full}
    \mathcal{F}_{\omega_0} \subset \mathcal{F}_{\omega_0,k}.
\end{equation}

\noindent \textbf{$k$ is far away from the origin point.} For example, $\vec{b}$ is initialized as $10$. 
Because the frequencies of each sub-functions are further increased for $\vec{b}=10$ (in Fig.~\ref{fig:sec_3.3}(b), the frequency is increased from the orange box to purple box), the supported frequency set $\mathcal{F}_{\omega_0,k}$ is increased a lot compared with the one in Eqn.~\ref{eqn:freq_set_small_b}, thus the Eqn.~\ref{eqn:siren_in_finer_full} also holds when $k$ is far away from the origin point.

Although the analysis is built upon the Sine function, similar relationship (Eqn.~\ref{eqn:siren_in_finer_full}) could also be obtained when other activation functions are used. In summary, 
\begin{prop}
    The supported frequency set $\mathcal{F}_{p,k}$ of FINER++ increases with the increase of the initialization range of $\vec{b}$, and the supported ``frequency'' set $\mathcal{F}_{p}$ in previous INRs is a subset of $\mathcal{F}_{\omega_0,k}$ in FINER++.
\end{prop}


\subsection{Neural Tangent Kernel Perspective}
Neural tangent kernel (NTK) theory~\cite{jacot2018neural} views the training of neural network as kernel regression, where the convergence of the network on fitting signals could be derived by analysing the diagonal property of the NTK or the distribution of NTK's eigenvalues. Generally speaking, stronger diagonal property results in better shift-invariance and better convergence, more larger eigenvalues leads to faster convergence for high-frequency components~\cite{tancik2020fourier,bai2023physics}.

Without loss of generality, we focus on a simple case, \textit{i.e.}, the signal to be learned has 1D input and 1D output and FINER++ has 1 hidden layer with $n$ neurons, such a network could be written as $f(x;\theta)=\sum_{k=1}^{n}c_{k}\sigma(w_{k}x+b_{k})$, where $\sigma(x)=\sin ((|x|+1)x)$ is the activation function. According to the definition, the NTK of FINER++ could be calculated as\footnote{Please refer the supplemental material for details of derivation.},
\begin{equation}
\begin{aligned}
        &\mathbf{K}(x_i,x_j)
        =\mathbb{E}_{\theta}\left< \nabla_{\theta}f(x_i;\theta), \nabla_{\theta}f(x_j;\theta) \right>\\
        =&(x_ix_j+1)\mathbb{E}_{\theta}\sum_{k=1}^{n}c_{k}^{2}\underbrace{(2|g_{k}(x_i)|+1)(2|g_{k}(x_j)|+1)}_{\text{Scale term}}\\
        &\underbrace{\cos{((|g_{k}(x_i)|+1)g_{k}(x_i))}\cos{((|g_{k}(x_j)|+1)g_{k}(x_j))}}_{\text{Sign term}}\\
        &\text{where}\:\: g_{k}(x_i)=w_{k}x_i+b_k.
\end{aligned}
\label{eqn:NTK_cal}
\end{equation}
It is observed that, the scale term is approximately proportional to the absolution of bias $b_k$ for all nodes of the kernel, however the change rule of sign term for diagonal elements differs significantly from non-diagonal elements. Specifically, the sign term is always a positive value for diagonal elements while could be either positive or negative for non-diagonal elements.

As a result, the diagonal elements $\mathbf{K}(x_i,x_i)$ are increased with the increase of $|b|$, while the non-diagonal elements $\mathbf{K}(x_i,x_j)$ can be very small, appearing as a diagonal enhanced kernel. According to \cite{jacot2018neural,tancik2020fourier}, NTK with a stronger diagonal property provides better shift-invariance, \textit{i.e.}, the coordinates in the training set are little coupled with each other during the training process, thus the signal could be better learned.

\begin{figure}[!t]
    \begin{center}
        \includegraphics[width=0.99\linewidth]{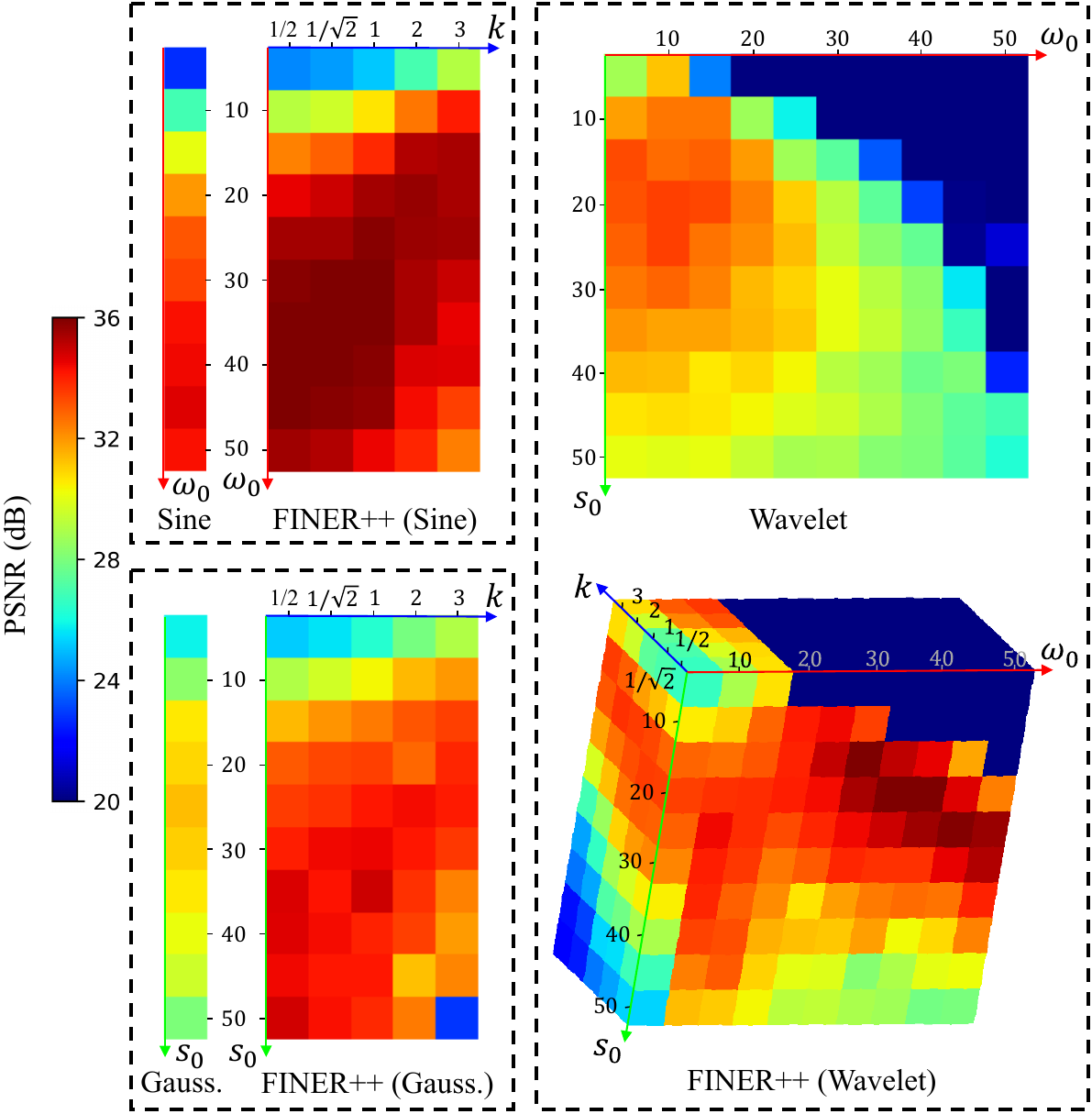}
        \caption{Enhanced robustness of FINER++ in parameter selection. The image demonstrates the performance of 2D image representation between previous INRs and their respective FINER++ versions across various parameters. FINER++ surpasses previous INRs not only by achieving the highest PSNR value but also by consistently outperforming them across a broad range of parameter selections.}
        \label{fig:enhanced_cap}
    \end{center}
\end{figure} 

\begin{figure*}[t] 
    \begin{center}
        \includegraphics[width=\linewidth]{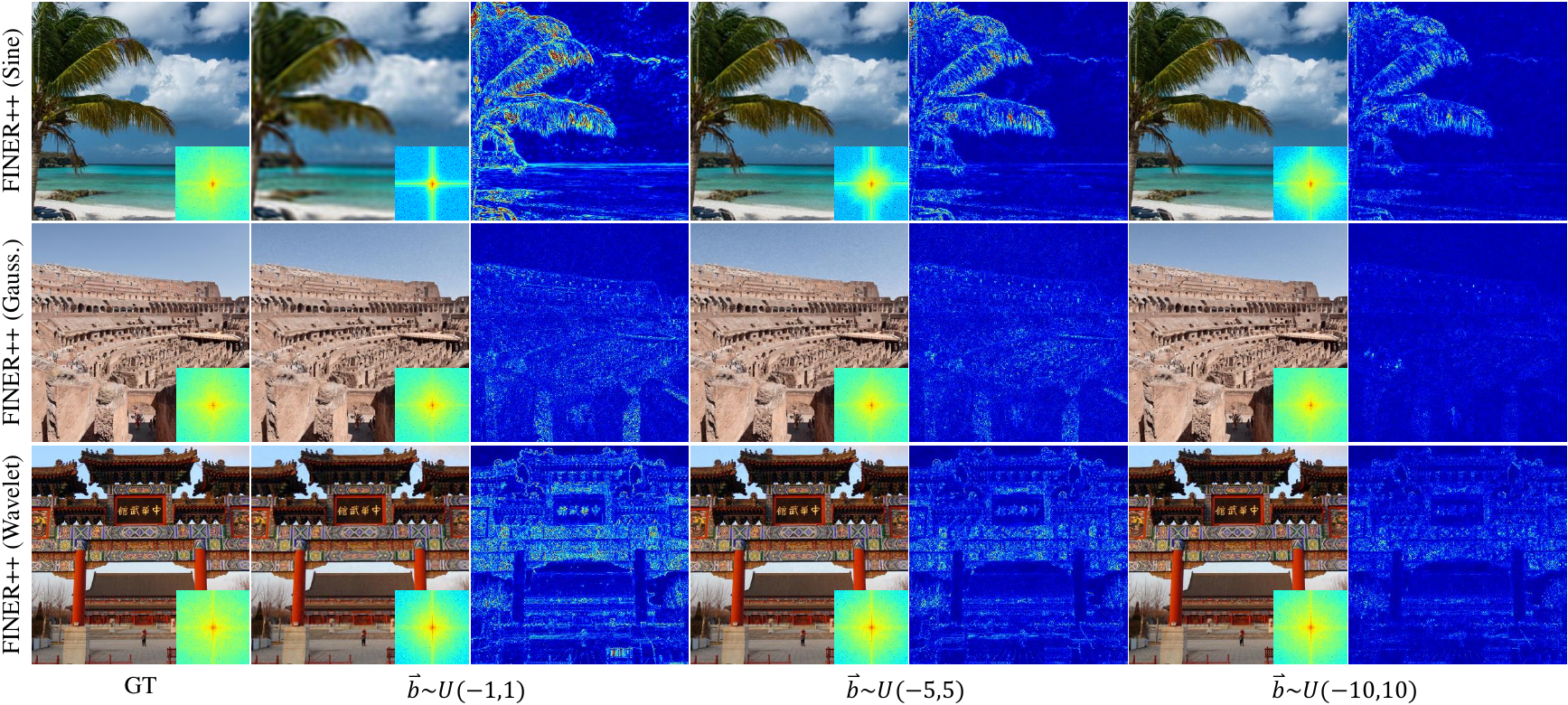}
        \caption{Comparisons of FINER++'s behaviors with different initializations applied to bias vector $\vec{b}$. For each image, the right-bottom box visualizes its Fourier spectrum. With an increased initialization range, more high-``frequency'' content can be represented, leading to a significant reduction in error.}
        \label{fig:finer_behavior_vis}
    \end{center}
\end{figure*}

In Figs.~\ref{fig:NTK_vis}, the left 4 sub-figures in the first row visualize the NTKs of FINER++ with Sine activation for learning a 1D signal with 1024 coordinates. It is observed that the diagonal property of NTK is enhanced with the increase of initialization range of $b$, verifying the analysis mentioned above. The right-most sub-figure visualizes the changes of eigenvalue distribution, it is observed that the number of eigenvalues which are larger than $10^0$ is significantly increased when larger initialization range is applied to bias.

Although the analysis above is based on the Sine activation, similar observations hold true when using Gaussian and Wavelet activation functions. In Fig.~\ref{fig:NTK_vis}, the second and third rows depict the Neural Tangent Kernels (NTKs) and the distribution of corresponding eigenvalues for FINER++ with Gaussian and Wavelet activation functions, respectively. It is observed that the diagonal property of the NTKs is strengthened, and the number of large eigenvalues (e.g., greater than $10^0$) increases with the widening of the initialization range of $\vec{b}$. It's worth noting that, due to the nonlinear scaling of parameters in Gaussian and Wavelet functions with respect to frequency, some off-diagonal elements in the NTKs of FINER++ (Gauss.) and FINER++ (Wave) are also amplified.

\begin{figure}[t] 
    \begin{center}
        \includegraphics[width=\linewidth]{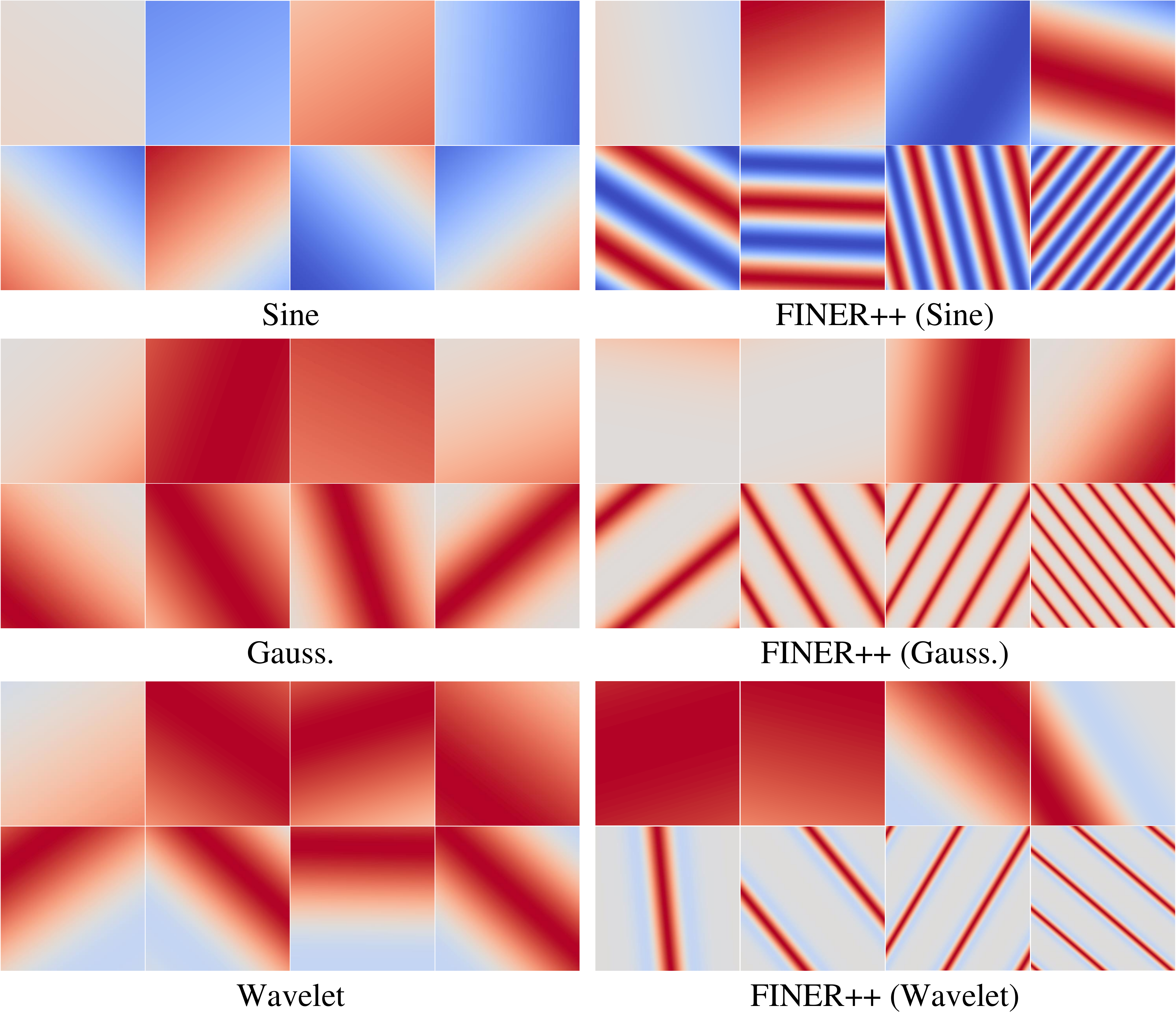}
        \caption{Comparisons of the first layer outputs between SIREN and FINER. For each method, the first row demonstrates 4 neurons with smallest frequencies of the first layer, and the second row refers to 4 neurons with largest frequencies.}
        \label{fig:neurons_cmp}
    \end{center}
\end{figure}

\subsection{Discussion of parameters selection}
Compared with previous INRs, FINER++ introduces an additional parameter $k$, which represents the initialization range of the bias vector. This added parameter not only addresses the spectral bias and capacity-convergence gap but also enhances the robustness of parameter selection. Fig.~\ref{fig:enhanced_cap} compares image representations between FINER++ and previous INRs across various parameters. All six INRs share the same network configuration: 3 layers with 256 neurons per layer. The comparison shows that FINER++ versions (Sine, Gauss., and Wavelet) achieve higher PSNR values compared to their previous counterparts. Moreover, FINER++ consistently outperforms previous INRs over a wide range of parameter settings, demonstrating enhanced robustness in parameter selection.


\definecolor{red}{RGB}{244,152,178}
\definecolor{orange}{RGB}{251,211,210}
\definecolor{yellow}{RGB}{253,238,238}

\begin{table*}
\centering
\caption{Quantitative comparisons on image fitting. We color code each cell as \colorbox{red}{best}, \colorbox{orange}{second best}, and \colorbox{yellow}{third best}.}
\begin{tabular}{lccccccc}
\toprule
 {Metrics}
& {PEMLP} & {SIREN} & {FINER++ (Sine)} & {Gaussian} & {FINER++ (Gauss.)} & {WIRE} & {FINER++ (Wavelet)} \\
\midrule
PSNR $\uparrow$     & 29.60      & 38.52                        & \cellcolor{red}40.76 (2.24$\uparrow$)    & 35.39    & \cellcolor{yellow}38.60 (3.21$\uparrow$)      & 31.31     & \cellcolor{orange}39.90 (8.59$\uparrow$)\\
SSIM $\uparrow$     & 0.8484     & \cellcolor{yellow}0.9724      & \cellcolor{red}0.9790 (0.0066$\uparrow$)   & 0.9455   & 0.9672 (0.0217$\uparrow$)                      & 0.8738    & \cellcolor{orange}0.9781 (0.1043$\uparrow$)\\
LPIPS $\downarrow$  & 1.21e-1    & \cellcolor{yellow}5.52e-3    & \cellcolor{red}2.56e-3 (2.96e-3$\downarrow$)  & 1.91e-2  & 7.90e-3 (5.99e-3$\downarrow$)                      & 6.45e-2   & \cellcolor{orange}5.00e-3 (1.45$\downarrow$)\\
\bottomrule
\end{tabular}
\label{tab:res_img}
\end{table*}


\begin{figure*} 
    \begin{center}
        \includegraphics[width=\linewidth]{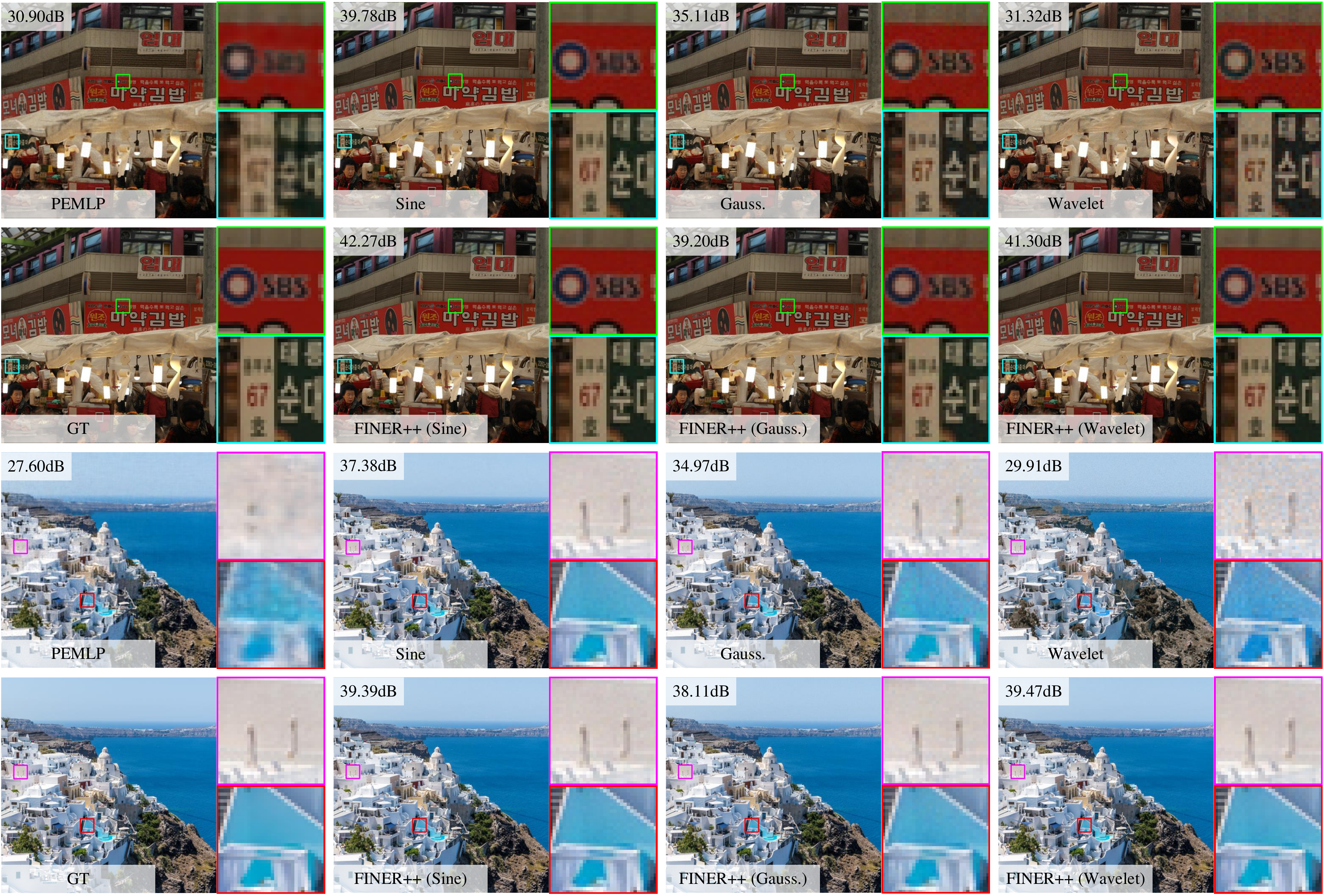}
        \caption{Qualitative comparisons between the FINER++ and baselines on fitting images.}
        \label{fig:res_img}
    \end{center}
\end{figure*}

\section{EXPERIMENTS}
We apply FINER++ to four tasks to verify the behaviors and performance, including 2D image fitting, 3D signed distance field representation, 5D neural radiance fields optimization and streamable INR transmission. 
\subsection{2D Image Fitting}
For the task of 2D image representation, the INR aims at learning a 2D function $f: \mathbb{R}^{2} \rightarrow \mathbb{R}^{3}$ with 2D pixel location input and 3D RGB color output, the loss function is designed as the $L_2$ distance between the network output and the ground truth. To evaluate the performance of INR, the widely used natural dataset~\cite{tancik2020fourier} which contains 16 images with $512\times 512$ resolution is adopted.

\subsubsection{FINER++'s behaviors under different initializations}
\label{sec:img_diff_bias}
To better understand the behavior of FINER++ under different initialization ranges, we set the scale parameters in previous INRs to small values ($\omega_0=1$ for Sine, $s_0=2.5$ for Gauss., and ${s_0,\omega_0}={2.5,5}$ for Wavelet) and vary the initialization ranges of bias vectors. According to the analysis in Sec.~\ref{sec:finer++_frame}, the supported ``frequency'' set of FINER++ increases with the initialization range of $\vec{b}$. Different curves in Fig.~\ref{fig:res_behavior_curves} reflect this behavior. The error and the spectrum maps of the learned images using FINER++ in Fig.~\ref{fig:finer_behavior_vis} also demonstrate this behavior qualitatively. For example, in the results of FINER++ with Sine activation (first row in Fig.~\ref{fig:finer_behavior_vis}), most of the energies are gathered at the low-frequency areas when $\vec{b}\sim \mathcal{U}(-1,1)$, resulting in blurred boundaries in reconstructed leaves. By increasing the initialization range, more high-frequency contents appear. When the Gaussian or Wavelet activation functions are used, the effects of initialization range for tuning FINER++'s supported frequency sets still work. Although the changes could not be well observed in the spectrum maps since the scale parameters here are not linearly corrected with the frequency, the error maps help recognizing the efficacy of changing initialization ranges of $\vec{b}$.

\begin{table*}
    \centering
    \caption{Quantitative comparisons on representing signed distance field. 
}
    \begin{tabular}{llccccccc}
    \toprule
        {Metrics} & {Scene} & {PEMLP} & {SIREN} & {FINER++ (Sine)} & {Gaussian} & {FINER++ (Gauss.)} & {WIRE} & {FINER++ (Wavelet)} \\
        \midrule
\multirow{6}*{Chamfer $\downarrow$} 
& Armadillo &3.559e-6 & 3.505e-6 & 3.348e-6 (1.57e-7$\downarrow$) & 1.778e-5 & \cellcolor{red}3.210e-6 (1.457e-6$\downarrow$) & \cellcolor{yellow}3.346e-6 & \cellcolor{orange}3.291e-6 (5.50e-8$\downarrow$) \\
 & Dragon & \cellcolor{orange}2.081e-6 & 2.759e-6 & 2.364e-6 (3.95e-7$\downarrow$) & 7.427e-6 & 2.172e-6 (5.255e-6$\downarrow$) & \cellcolor{yellow}2.101e-6 & \cellcolor{red}1.959e-6 (1.42e-7$\downarrow$)\\
 & Lucy & 2.224e-6 & 2.493e-6 & \cellcolor{orange}2.119e-6 (3.74e-7$\downarrow$) & 5.494e-6 & \cellcolor{yellow}2.176e-6 (3.318e-6$\downarrow$) & 2.238e-6 & \cellcolor{red}2.032e-6 (2.06e-7$\downarrow$)\\
 & Thai Statue & 5.284e-6 & 4.481e-6 & \cellcolor{yellow}3.580e-6 (9.01e-7$\downarrow$) & 1.618e-5 & \cellcolor{orange}3.190e-6 (1.299e-5$\downarrow$) & 3.979e-6 & \cellcolor{red}3.072e-6 (9.07e-7$\downarrow$)\\
 & Avg. & 3.441e-6 & 3.438e-6 & \cellcolor{yellow}3.087e-6 (3.51e-7$\downarrow$) & 1.262e-5 & \cellcolor{orange}2.898e-6 (9.722e-6$\downarrow$) & 3.252e-6 & \cellcolor{red}2.483e-6 (7.69e-7$\downarrow$)\\
 \midrule
 \multirow{6}*{IOU $\uparrow$} 
 & Armadillo & 9.870e-1 & 9.895e-1 & \cellcolor{yellow}9.899e-1 (4.00e-4$\uparrow$) & 9.768e-1 & \cellcolor{red}9.919e-1 (1.51e-2$\uparrow$) & 9.893e-1 & \cellcolor{orange}9.901e-1 (8.00e-4$\uparrow$) \\
 & Dragon & \cellcolor{yellow}9.750e-1 & 9.666e-1 & 9.725e-1 (5.90e-3$\uparrow$) & 9.679e-1 & \cellcolor{orange}9.771e-1 (9.20e-3$\uparrow$) & 9.723e-1 & \cellcolor{red}9.806e-1 (8.30e-3$\uparrow$) \\
 & Lucy & \cellcolor{orange}9.760e-1 & 9.721e-1 & \cellcolor{yellow}9.756e-1 (3.50e-3$\uparrow$) & 9.601e-1 & 9.733e-1 (1.32e-2$\uparrow$) & 9.707e-1 & \cellcolor{red}9.807e-1 (1.00e-2$\uparrow$)\\
 & Thai Statue & 9.526e-1 & 9.514e-1 & \cellcolor{orange}9.625e-1 (1.11e-2$\uparrow$) & 9.481e-1 & \cellcolor{red}9.647e-1 (1.66e-2$\uparrow$) & 9.565e-1 & \cellcolor{yellow}9.623e-1 (5.80e-3$\uparrow$)\\
 & Avg. & 9.769e-1 & 9.749e-1 & \cellcolor{yellow}9.790e-1 (4.10e-3$\uparrow$) & 9.692e-1 & \cellcolor{orange}9.806e-1 (1.14e-2$\uparrow$) & 9.760e-1 & \cellcolor{red}9.817e-1 (5.70e-3$\uparrow$)\\
 \bottomrule
    \end{tabular}
    \label{tab:res_sdf}
\end{table*}



According to \cite{yuce2022structured}, the first layer of INR plays the role of frequency encoding. We visualize 8 neurons output from total 256 neurons in first layer from previous INRs and their FINER++'s versions, where 4 neurons in the 1st row have the smallest frequencies and the last 4 neurons have the largest frequencies (see Fig.~\ref{fig:neurons_cmp}). It is observed that different neurons in previous INRs encode similar frequencies, resulting in a waste of neurons. On the contrary, different neurons in FINER++ focus on different frequencies, therefore, better representational ability is achieved in FINER++.

\begin{figure*}[!t] 
    \begin{center}
        \includegraphics[width=\linewidth]{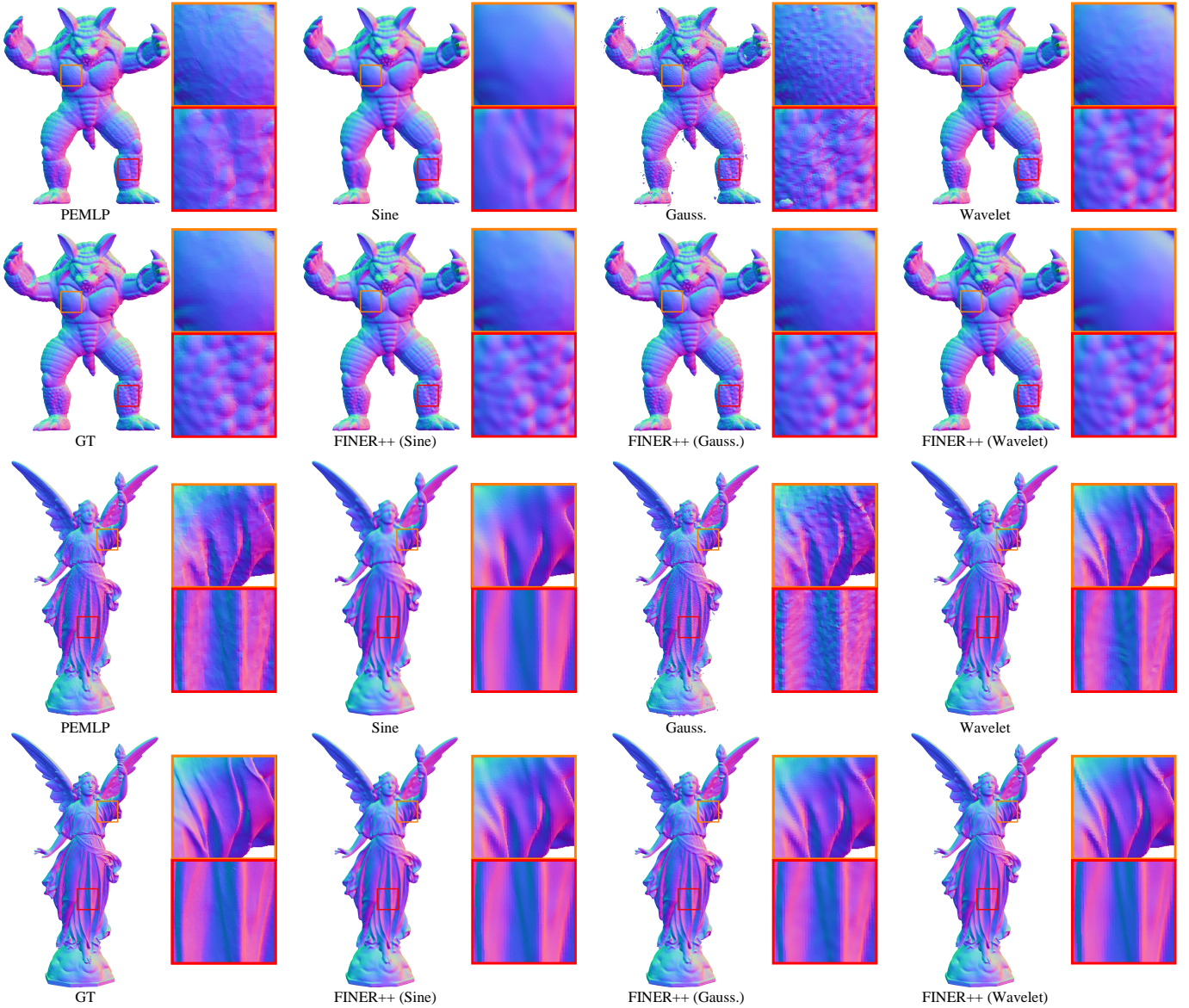}
        \caption{Qualitative comparisons on representing the signed distance fields of Armadillo and Lucy.}
        \label{fig:res_sdf}
    \end{center}
\end{figure*} 

\begin{table*}
    \centering
    \caption{Quantitative comparisons on novel view synthesis.}
    \begin{tabular}{{llccccccc}}
\toprule
    {Metrics} & {Scene} & {PEMLP} & {SIREN} & {FINER++ (Sine)} & {Gaussian} & {FINER++ (Gauss.)} & {WIRE} & {FINER++ (Wavelet)} \\
\midrule
\multirow{9}*{PSNR $\uparrow$}
 & Chair & 31.32 & 33.31 & \cellcolor{red}33.90 (0.59$\uparrow$) & 32.68 & \cellcolor{yellow}33.49 (0.81$\uparrow$) & 29.31 & \cellcolor{orange}33.78 (4.47$\uparrow$)\\
 & Drums & 20.18 & \cellcolor{orange}24.89 & \cellcolor{red}24.90 (0.01$\uparrow$) & 23.16 & 24.55 (1.39$\uparrow$)& 22.22 & \cellcolor{orange}24.89 (2.67$\uparrow$)\\
 & Ficus & 24.49 & 27.26 & \cellcolor{orange}28.70 (1.44$\uparrow$) & 26.10 & \cellcolor{yellow}28.41 (2.31$\uparrow$) & 25.91 & \cellcolor{red}28.81 (2.90$\uparrow$)\\
 & Hotdog & 30.59 & \cellcolor{yellow}32.85 & \cellcolor{red}33.05 (0.20$\uparrow$)& 32.17 & 32.28 (0.11$\uparrow$) & 30.11 & \cellcolor{orange}32.90 (2.79$\uparrow$)\\
 & Lego & 25.90 & \cellcolor{yellow}29.60 & \cellcolor{red}30.04 (0.44$\uparrow$)& 28.29 & 29.35 (1.06$\uparrow$) & 25.76 & \cellcolor{orange}29.78 (4.02$\uparrow$)\\
 & Materials & 25.16 & \cellcolor{orange}27.13 & \cellcolor{yellow}27.05 (-0.08$\uparrow$) & 26.19 & 26.91 (0.72$\uparrow$) & 25.05 & \cellcolor{red}27.21 (2.16$\uparrow$)\\
 & Mic & 26.38 & 33.28 & \cellcolor{orange}33.96 (0.68$\uparrow$)& 33.59 & \cellcolor{yellow}33.75 (0.16$\uparrow$) & 32.35 & \cellcolor{red}34.36 (2.01$\uparrow$)\\
 & Ship & 21.46 & 22.25 & \cellcolor{orange}22.47 (0.22$\uparrow$) & 22.28 & \cellcolor{yellow}22.44 (0.16$\uparrow$) & 21.15 & \cellcolor{red}22.66 (1.51$\uparrow$)\\
 & Avg. & 25.69 & 28.82 & \cellcolor{orange}29.26 (0.44$\uparrow$)& 28.06 & \cellcolor{yellow}28.90 (0.84$\uparrow$) & 26.48 & \cellcolor{red}29.30 (2.82$\uparrow$)\\
\midrule
\multirow{9}*{SSIM $\uparrow$}
& Chair & 0.960 & 0.971 & \cellcolor{orange}0.973 (0.002$\uparrow$) & 0.967 & \cellcolor{yellow}0.972 (0.005$\uparrow$) & 0.938 & \cellcolor{red}0.975 (0.037$\uparrow$)\\
& Drums & 0.814 & \cellcolor{orange}0.912 & \cellcolor{yellow}0.911 (-0.001$\uparrow$)& 0.883 & 0.909 (0.026$\uparrow$) & 0.858 & \cellcolor{red}0.917 (0.059$\uparrow$)\\
& Ficus & 0.914 & 0.947 & \cellcolor{yellow}0.958 (0.011$\uparrow$)& 0.933 & \cellcolor{orange}0.960 (0.027$\uparrow$) & 0.931 & \cellcolor{red}0.963 (0.032$\uparrow$)\\
& Hotdog & 0.945 & \cellcolor{orange}0.960 & \cellcolor{yellow}0.959 (-0.001$\uparrow$)& 0.956 & 0.953 (-0.003$\uparrow$) & 0.938 & \cellcolor{red}0.963 (0.025$\uparrow$)\\
& Lego & 0.904 & \cellcolor{yellow}0.948 & \cellcolor{red}0.951 (0.003$\uparrow$)& 0.932 & 0.945 (0.013$\uparrow$) & 0.886 & \cellcolor{red}0.951 (0.065$\uparrow$)\\
& Materials & 0.909 & \cellcolor{red}0.932 & \cellcolor{yellow}0.928 (-0.004$\uparrow$)& 0.916 & \cellcolor{yellow}0.928 (0.012$\uparrow$) & 0.898 & \cellcolor{orange}0.931 (0.033$\uparrow$)\\
& Mic & 0.960 & 0.979 & \cellcolor{yellow}0.981 (0.002$\uparrow$)& 0.980 & \cellcolor{orange}0.982 (0.002$\uparrow$) & 0.978 & \cellcolor{red}0.983 (0.005$\uparrow$)\\
& Ship & 0.754 & 0.788 & \cellcolor{orange}0.792 (0.004$\uparrow$)& 0.782 & \cellcolor{yellow}0.790 (0.008$\uparrow$) & 0.734 & \cellcolor{red}0.796 (0.062$\uparrow$)\\
& Avg. & 0.895 & \cellcolor{yellow}0.930 & \cellcolor{orange}0.932 (0.002$\uparrow$)& 0.919 & \cellcolor{yellow}0.930 (0.011$\uparrow$) & 0.895 & \cellcolor{red}0.935 (0.004$\uparrow$)\\
\midrule
\multirow{9}*{LPIPS $\downarrow$}
 & Chair & 0.026 & 0.017 & \cellcolor{orange}0.015 (0.002$\downarrow$) & 0.019 & \cellcolor{yellow}0.016 (0.003$\downarrow$)& 0.035 & \cellcolor{red}0.014 (0.021$\downarrow$)\\
 & Drums & 0.185 & \cellcolor{orange}0.051 & \cellcolor{yellow}0.052 (-0.001$\downarrow$) & 0.082 & 0.055 (0.027$\downarrow$)& 0.106 & \cellcolor{red}0.049 (0.057$\downarrow$)\\
 & Ficus & 0.056 & 0.033 & \cellcolor{orange}0.024 (0.009$\downarrow$) & 0.055 & \cellcolor{orange}0.024 (0.031$\downarrow$)& 0.045 & \cellcolor{red}0.022 (0.023$\downarrow$)\\
 & Hotdog & 0.037 & \cellcolor{orange}0.032 & 0.033 (-0.001$\downarrow$) & \cellcolor{orange}0.032 & 0.038 (-0.006$\downarrow$)& 0.048 & \cellcolor{red}0.028 (0.020$\downarrow$)\\
 & Lego & 0.070 & 0.030 & \cellcolor{red}0.025  (0.005$\downarrow$) & 0.040 & \cellcolor{yellow}0.029 (0.011$\downarrow$)& 0.075 & \cellcolor{orange}0.027 (0.048$\downarrow$)\\
 & Materials & 0.050 & \cellcolor{red}0.028 & \cellcolor{yellow}0.032  (-0.004$\downarrow$) & 0.037 & 0.033 (0.004$\downarrow$)& 0.058 & \cellcolor{orange}0.029 (0.029$\downarrow$)\\
 & Mic & 0.057 & 0.014 & \cellcolor{red}0.010 (0.004$\downarrow$) & 0.013 & \cellcolor{orange}0.012 (0.001$\downarrow$)& 0.014 & \cellcolor{red}0.010 (0.004$\downarrow$)\\
 & Ship & 0.190 & 0.116 & \cellcolor{orange}0.108  (0.008$\downarrow$) & 0.136 & \cellcolor{orange}0.108 (0.028$\downarrow$)& 0.172 & \cellcolor{red}0.100 (0.072$\downarrow$)\\
 & Avg. & 0.084 & 0.040 & \cellcolor{orange}0.037 (0.003$\downarrow$) & 0.052 & \cellcolor{yellow}0.039 (0.012$\downarrow$)& 0.069 & \cellcolor{red}0.035 (0.034$\downarrow$)\\
\bottomrule
    \end{tabular}
    \label{tab:res_nerf}
\end{table*}

\begin{figure*}[!t] 
    \begin{center}
        \includegraphics[width=0.95\linewidth]{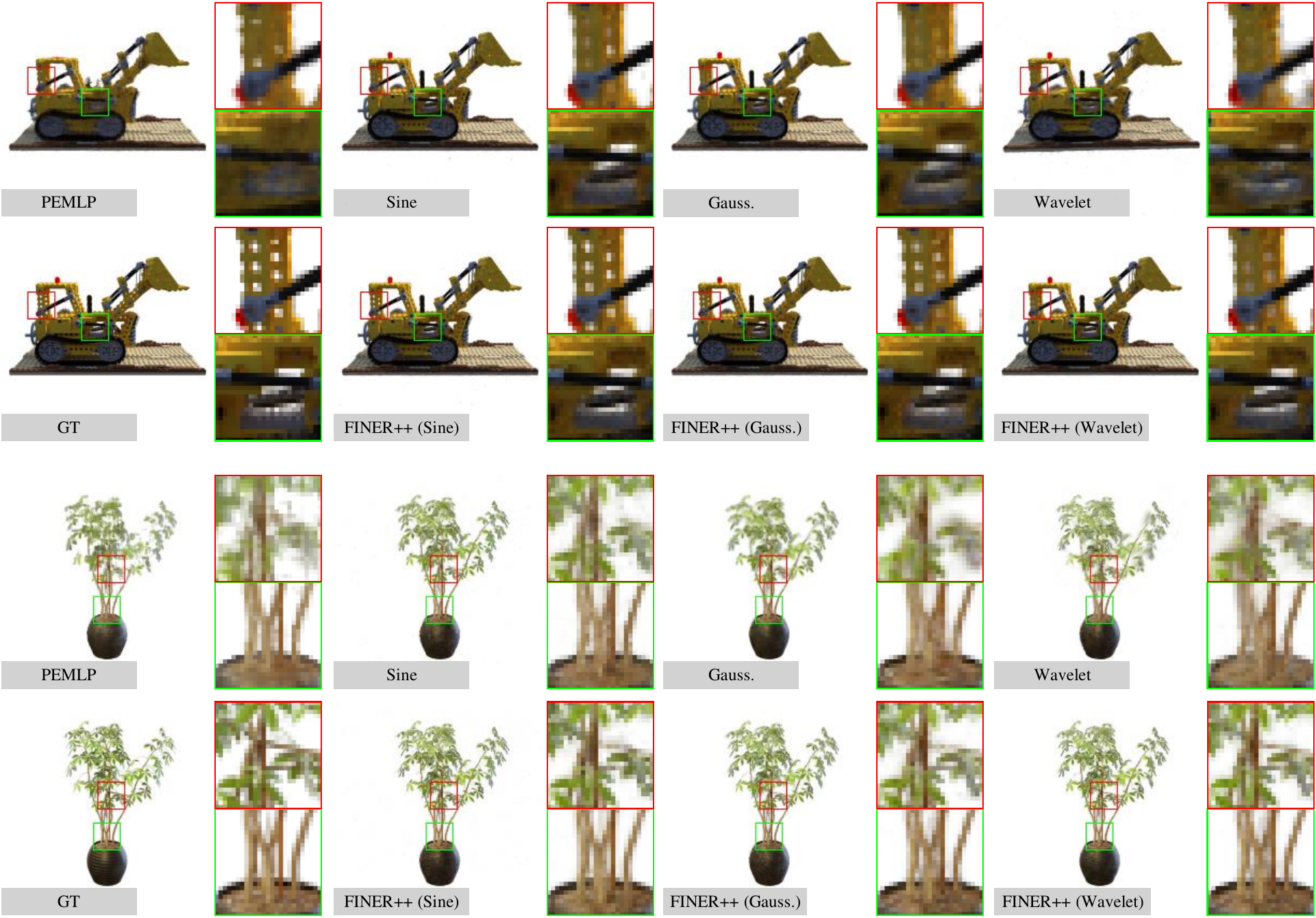}
        \caption{Qualitative comparisons between the FINER++ and baselines on NeRF.}
        \label{fig:res_nerf}
    \end{center}
\end{figure*}

\subsubsection{Comparisons with the State-of-the-arts}
We compare FINER++ with four classical INRs, \textit{i.e.}, the Fourier feature embedding (PEMLP)~\cite{mildenhall2020nerf}, INRs with periodic activation functions (SIREN)~\cite{sitzmann2020implicit}, Gaussian activation functions (Gauss.)~\cite{ramasinghe2022beyond} and wavelet activation functions (WIRE)~\cite{saragadam2023wire}. Note that the latter three methods are all combined with the FINER++ framework to better evaluate the performance. For a fair comparison, all INRs are set with a same network configuration (3 hidden layers with 256 neurons per layer, which is also a common configuration in literature~\cite{saragadam2023wire}) and are trained with the same Adam optimizer~\cite{kingma2015adam} and $L_2$ loss function between the network output and the ground truth, other parameters are set according to the open-source codes released by authors. In FINER++, $k$ is set as $\frac{1}{\sqrt{2}}$, $1$ and $1$ for Sine, Gauss. and Wavelet backbones, respectively. Tab.~\ref{tab:res_img} compares FINER++ with others quantitatively. FINER++ outperforms other methods in all three metrics and there are significantly improvements compared with backbone INRs.  Fig.~\ref{fig:res_img} demonstrates the details of different methods. By extending previous INRs to their FINER++'s versions, over-smoothed text boundaries becomes clear (Sine backbone, \textit{e.g.}, the texts '67' and 'SBS' in the cyan and green boxes, respectively) and unwelcome serious artefacts in the smooth background are removed (Gauss. and Wavelet backbones, \textit{e.g.}, white and red billboards in the cyan and green boxes, respectively, as well as the while wall and the blue glass in the purple and red boxes, respectively.)

\subsection{3D Shape Representation}
Signed distance field (SDF) is one of the most commonly used implicit surface representations in the computer graphics~\cite{jones2006distance}. As the name implies, SDF characterizes the distance from the given 3D point to the closest surface using a continuous function, and the sign of the distance is used to denote whether the point is inside (negative) or outside (positive) the surface. Recently, representing the SDF using INR is drawing more and more attentions~\cite{wang2021neus,li2023neuralangelo}. Given a 3D point $\vec{x}$, INR learns a 3D mapping function $f: \mathbb{R}^{3} \rightarrow \mathbb{R}^{1}$ to output the signed distance field values $s$. 
 We apply the FINER++ to this task and compare to four classical INRs mentioned above. In this task, $k$ is set as $1$, $1$ and $\frac{1}{\sqrt{2}}$ for Sine, Gauss. and Wavelet backbones, respectively. 
In the experiment, 4 shapes from public dataset~\cite{standord-3D-scanning} 
 are used for evaluation. For a fair comparison, all methods use a same network configuration, \textit{i.e.}, 3 layers with 256 neurons per layer and the same coarse-to-fine loss function is used according to \cite{lindell2021bacon}. In the training stage, 10k points are randomly sampled in each iteration and is repeated $200k$ iterations. In the testing stage, a $512^3$ grid is extracted for evaluation and visualization.

Tab.~\ref{tab:res_sdf} provides quantitative comparisons between the proposed FINER++ and four baselines. Because FINER++ provides more freedom for tuning the supported frequency set, the performance of all three baseline methods (SIREN, Gaussian and Wavelet) are improved when the FINER++ framework is applied, and the best results are achieved by the FINER++ with the Wavelet backbone. Fig.~\ref{fig:res_sdf} compares the reconstructed details visually on the Armadillo and Lucy rendered using Marching Cubes~\cite{lorensen1998marching}. In each scene, two representative regions are zoomed-in for comparisons, \textit{i.e.}, the low-frequency smooth pectoral and the high-frequency rough shank in Armadillo, as well as the mid-frequency wrinkles on clothes in underarm and hemline in Lucy. For PEMLP, because the pre-defined frequency may not match the frequency distribution in the SDF of Armadillo and Lucy, all of the pectoral, shank and wrinkles on clothes are not well represented. SIREN represents the smooth pectoral well but loses the details of the shank and over-smooths the wrinkles on clothes. Gaussian could not provide stable representation for all shapes (Tab.~\ref{tab:res_sdf}) and there are obvious artefacts in the reconstructed SDF such as the noise outside the shape in Fig.~\ref{fig:res_sdf}. WIRE overcomes the limitation of SIREN for representing high-frequency shank at the cost of rough pectoral and wrinkles on clothes. By applying the FINER++ framework to three baselines, all of the problems of high-frequency noise and over-smooth textures could be well addressed and consistent performance are provided in all the low-, mid- and high-frequency components.

\begin{figure*}[!t] 
    \begin{center}
        \includegraphics[width=\linewidth]{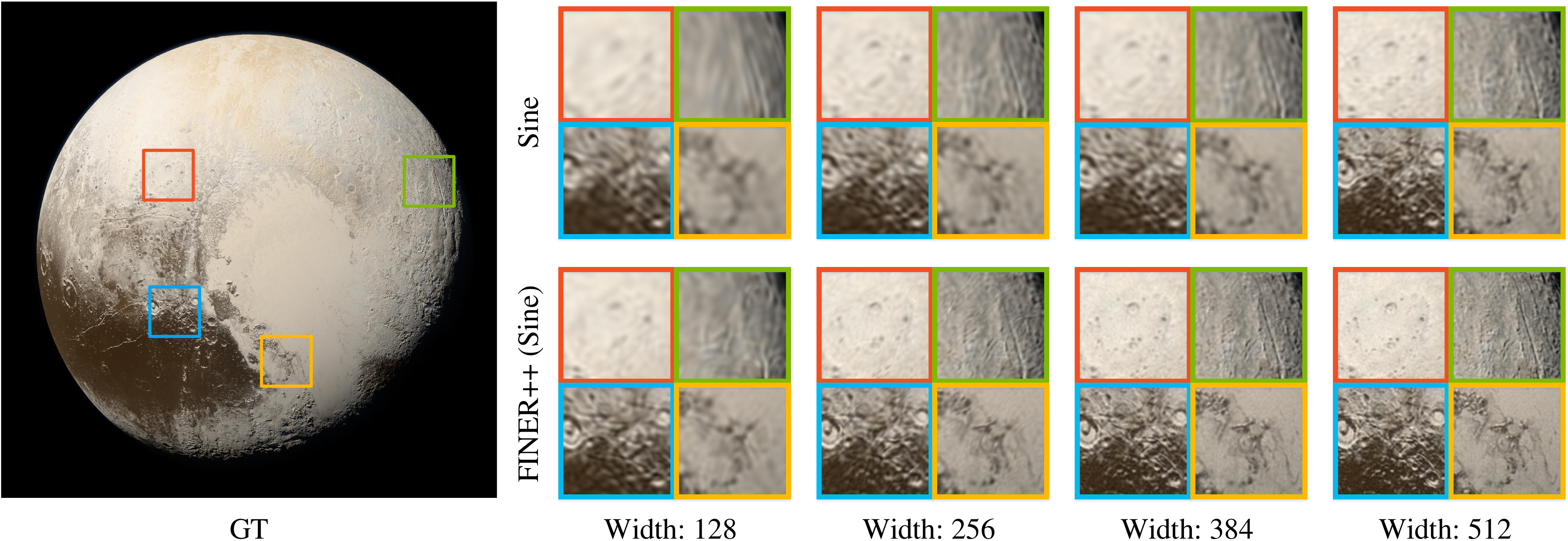}
        \caption{Comparisons on streamable INR transmission between SIREN and its FINER++ version on the gigapixel image 'Pluto'. FINER++ (Sine) outperforms the SIREN in all training stages (see Fig.~\ref{fig:pipeline_streamINR}) of streamable INR.}
        \label{fig:res_streamINR}
    \end{center}
\end{figure*}

\begin{figure}[t] 
    \begin{center}
        \includegraphics[width=\linewidth]{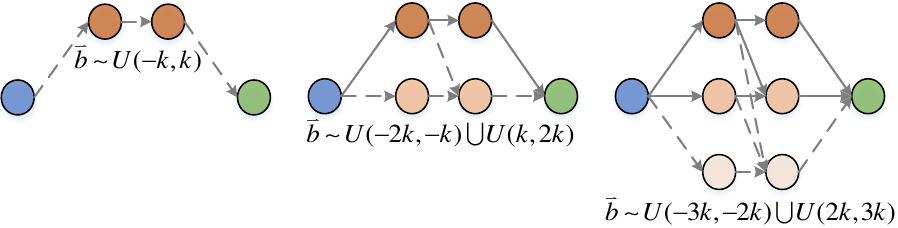}
        \caption{Pipeline of applying FINER++ to streamable INR transmission. From left to right, the initialization range of bias vector in widened network are increased to represent high-frequency components better.}
        \label{fig:pipeline_streamINR}
    \end{center}
\end{figure}

\subsection{Neural Radiance Fields Optimization}
Novel view synthesis, which aims at rendering realistic images at uncaptured poses from a set of images captured at different positions, is one of the key problems in both communities of computer vision and graphics. Recently, representing scenes as neural radiance fields (NeRF)~\cite{mildenhall2020nerf} using INR dominates this task due to the advantages of realism and scalability for embedding different rendering processes~\cite{tewari2022advances}. Given a 3D point 
$\vec{x}$ and a 2D observed direction $\vec{d}$, NeRF focuses on learning a 5D mapping function $f: \mathbb{R}^{5} \rightarrow \mathbb{R}^{4}$ with 5D coordinate $(\vec{x}, \vec{d})$ to its 3D color $c$ and 1D opacity $\sigma$. For any pixel $p$ in novel view images, its ray function in 3D space is firstly calculated using the in/extrinsic matrices of camera~\cite{hartley2003multiple}, then several points are sampled along the ray within a predefined depth range, furthermore the direction and position coordinates of these points are fed into the INR for querying the radiance value $(c,\sigma)$, finally the color $C(p)$ of the pixel is calculated using the differentiable volume rendering technique~\cite{max1995optical,mildenhall2020nerf},
\begin{equation}
\label{eqn:exp_nerf}
\begin{aligned}
C(p)&=\sum_{i=1}^N T_i\left(1-\exp \left(-\sigma_i \delta_i\right)\right) c_i \\
T_i&=\exp \left(-\sum_{j=1}^{i-1} \sigma_j \delta_j\right),
\end{aligned}
\end{equation}
where $\delta_i$ represents the distance between the neighbors in the sampled points.

We evaluate the FINER++ for this task and compare to four classical methods mentioned above ($k$ is set as $\frac{1}{\sqrt{3}}$, $\frac{1}{\sqrt{3}}$ and $1$ for Sine, Gauss. and Wavelet backbones, respectively). To better verify the advantages of FINER++ for representing high-frequency components, we follow the experimental setting of WIRE that only 25 images are used for training instead of commonly used 100 images. 
Tab.~\ref{tab:res_nerf} and Fig.~\ref{fig:res_nerf} provide quantitative and qualitative comparisons of FINER++ against different methods on the Blender dataset~\cite{mildenhall2020nerf}. FINER++ (Wavelet) achieves the best performance in almost all 8 scenes and all results of backbones are improved when the FINER++ is applied. Fig.~\ref{fig:res_nerf} demonstrates the advantage of FINER++ for representing high-frequency components visually. For example, the holes (red boxes) and the highlights (green boxes) in the frame of the truck are over-smoothed in the reconstructions of all baselines in Lego, however, these areas are all well reconstructed in the corresponding FINER++'s versions. These phenomenon also appears in the reconstructed results of Ficus, such as leaves (red boxes) and branches (green boxes).

\subsection{Streamable INR Transmission}
With the increasing use of INR in inverse optimization and signal representation, there is a growing demand to transfer reconstructed or represented INRs between different users. However, due to the strong coupling of INR parameters, challenges arise when distributing INRs to devices with varying resolutions or providing streaming services that require progressively decoding different signal components while ensuring consistency with transferred network parameters. To address these challenges, Cho et al.~\cite{cho2022streamable} propose a progressive training strategy. As depicted in Fig.~\ref{fig:pipeline_streamINR}, this strategy involves initially training a small, narrow network to model low-frequency components. Subsequently, the network width is increased, and the additional network parameters (represented by the dotted lines in Fig.~\ref{fig:pipeline_streamINR}) are trained to capture mid-frequency components, while the parameters from the initial step remain fixed (solid lines in Fig.~\ref{fig:pipeline_streamINR}). This process is repeated to model higher-frequency components. In their implementation, Cho et al. use a SIREN network with a default scale parameter ($\omega_0 = 30$) for both the initial narrow network and the subsequently added networks. However, as analysed in Sec.~\ref{sec:INR_properties}, SIREN exhibits a frequency-specified spectral bias, which limits its expressive power for modeling high-resolution signals under this training configuration.

FINER++ is inherently suitable for building streamable INR, especially in cases where the signal has high resolution. As analysed in Sec.~\ref{sec:finer++_frame}, the supported ``frequency'' set could be significantly manipulated by scaling the initialization range $[-k,k]$ of the bias vector. Consequently, the performance of FINER++-based streamable INR could be significantly enhanced by changing the initialization range of the bias vector in the widened network parameters. Fig.~\ref{fig:pipeline_streamINR} visualizes the pipeline of applying FINER++ in streamable INR transmission. Note that, the rule of setting initialization range (e.g., $\vec{b}\sim \mathcal{U}(-3k,2k)\cup\mathcal{U}(2k,3k)$) differs from the analysis in Sec.~\ref{sec:finer++_frame} to reduce the frequency overlap between neurons in different widths. Fig.~\ref{fig:res_streamINR} compares the streamable INR between the SIREN and FINER++ (Sine) in different stages qualitatively. It is observed that FINER++ (Sine) provides more clear details than SIREN not only in areas with different frequencies but also in all stages, demonstrating the advantages and potential of applying FINER++ in streamable INR.

\section{Conclusion}
We have proposed and verified the FINER++ which extends existing periodic/non-periodic functions to variable-periodic ones for activating INR. We have pointed out 3 characteristics of existing INRs, \textit{i.e.}, under-utilized definition domain in activation functions, ``frequency''-specified spectral bias and capacity-convergence gap. The proposed FINER++ addresses these characteristics by building a family of variable-periodic activation functions and initializing the bias vector to different ranges, where different sub-functions with different frequencies along the definition domain will be selected for activation. As a result, the latter two characteristics of existing INRs could be well handled. We have demonstrated the advantages of FINER++ over other INRs in image fitting, 3D shape representation, neural rendering and streamable INR transmission. In the future, we will focus on designing INRs without any ``frequency''-specified spectral bias.



%

\appendices




\ifCLASSOPTIONcaptionsoff
  \newpage
\fi



%



\bibliographystyle{IEEEtran}
\bibliography{IEEEabrv,./egbib}

%

\begin{IEEEbiography}[{\includegraphics[width=1in,height=1.25in,clip,keepaspectratio]{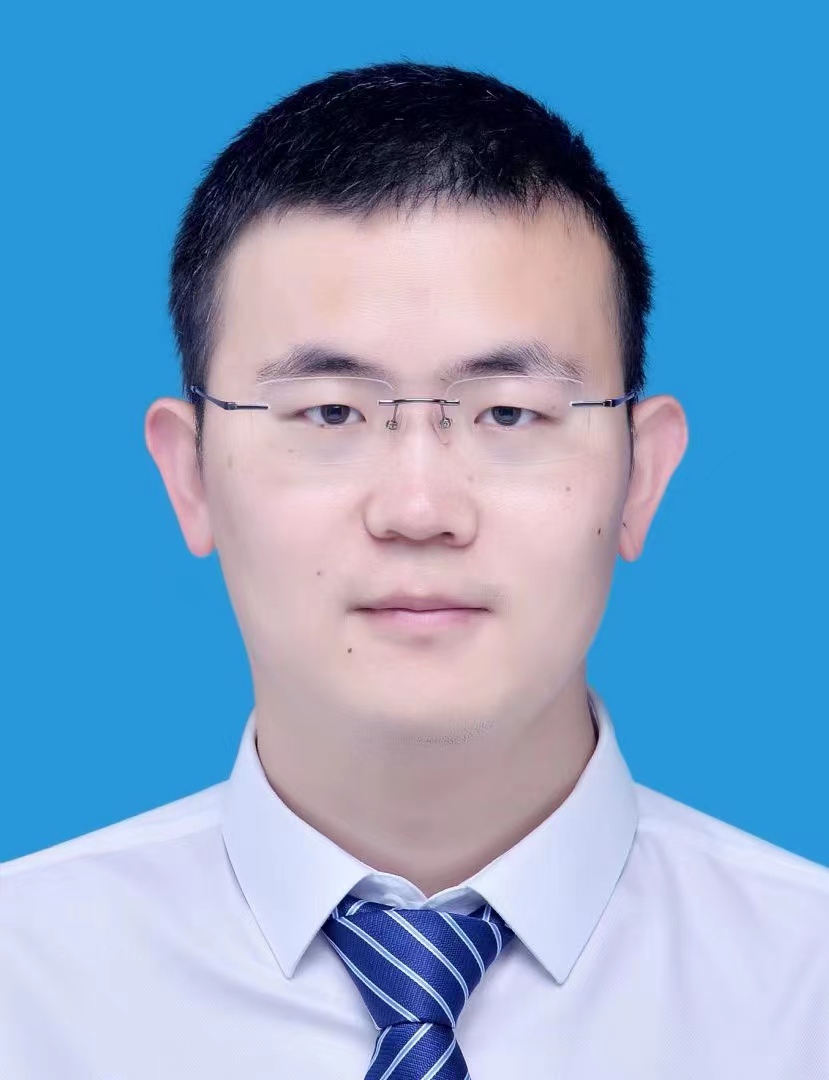}}]{Hao Zhu} is an Associate Researcher in the School of Electronic Science and Engineering, Nanjing University. He received the B.S. and Ph.D. degrees from Northwestern Polytechnical University in 2014 and 2020, respectively. He was a visiting scholar at the Australian National University. His research interests include computational photography and optimization for inverse problems.
\end{IEEEbiography}

\begin{IEEEbiography}[{\includegraphics[width=1in,height=1.25in,clip,keepaspectratio]{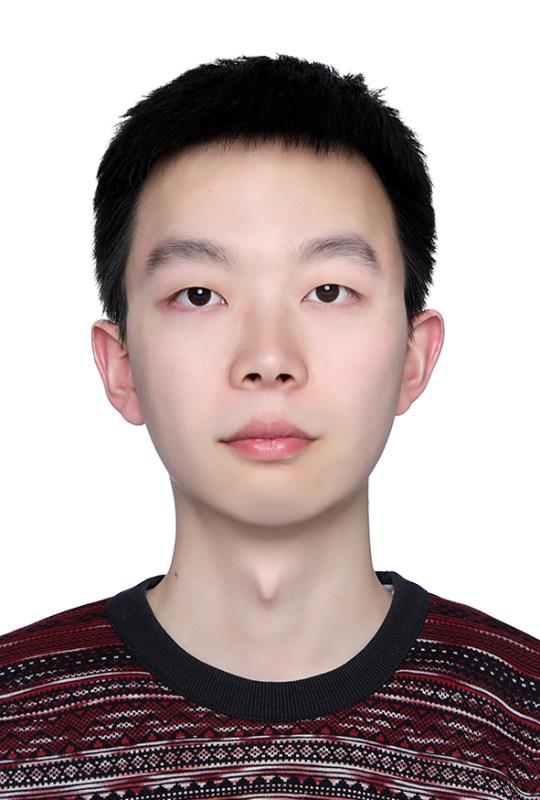}}]{Zhen Liu} is a graduate student in the Department of Computer Science and Technology, Nanjing University. He is co-supervised by Prof. Xun Cao and Prof. Yang Yu. He received his B.S. degree from the Beijing Institute of Technology in 2021. His research interests include computational photography and implicit neural representation.
\end{IEEEbiography}


\begin{IEEEbiography}[{\includegraphics[width=1in,height=1.25in,clip,keepaspectratio]{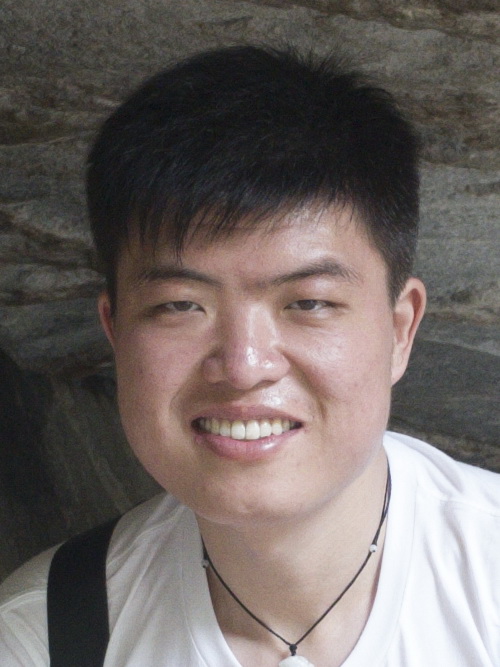}}]{Qi Zhang} is currently a lead researcher with Vivo. Before that, he was a researcher with Tencent AI Lab. He received his Ph.D. from the School of Computer Science at Northwestern Polytechnical University in 2021. He received CCF Doctorial Dissertation Award Nominee in 2021. His research interests include 3D vision, neural rendering, Gaussian Splatting, and AIGC.
\end{IEEEbiography}

\begin{IEEEbiography}[{\includegraphics[width=1in,height=1.25in,clip,keepaspectratio]{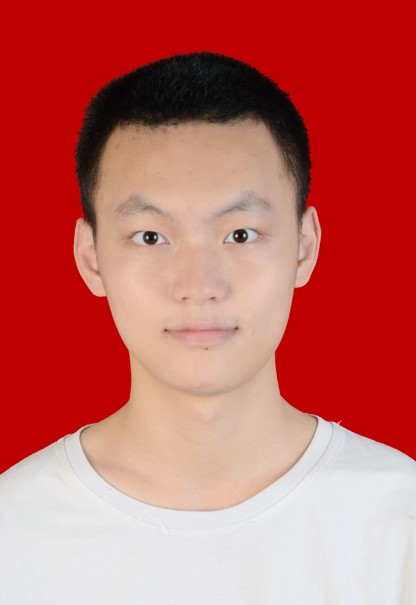}}]{Jingde Fu} received the B.S. degree in 2024 from the Department of Mathematics, Nanjing University, Nanjing, China. He is currently a graduate student for Ph.D degree in the Department of Mathematics, Nanjing University, Nanjing, China. His current research interests include numerical methods for partial differential equations and modeling methods driven by data and mechanism.
\end{IEEEbiography}

\begin{IEEEbiography}[{\includegraphics[width=1in,height=1.25in,clip,keepaspectratio]{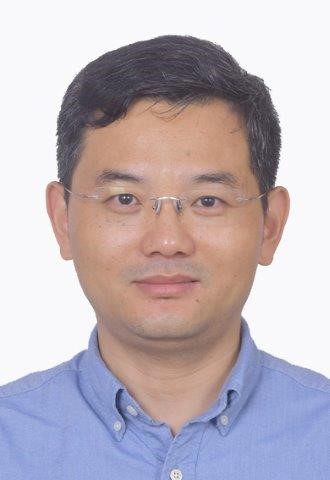}}]{Weibing Deng} received the B.S. degree in 1992, and the Ph.D. degree in 2002 from the Department of Mathematics, Nanjing University, Nanjing, China. He worked as a postdoctoral fellow in the Institute of Computational Mathematics, Chinese Academy of Sciences, form 2003 to 2004. He was a Visiting Scholar with the California Institute of Technology, USA, from 2007 to 2008. He is a Professor at the school of Mathematics, Nanjing University. His current research interests include analysis and computation of multi-scale problems, numerical homogenization methods, and modeling methods driven by data and mechanism.
\end{IEEEbiography}

\begin{IEEEbiography}[{\includegraphics[width=1in,height=1.25in,clip,keepaspectratio]{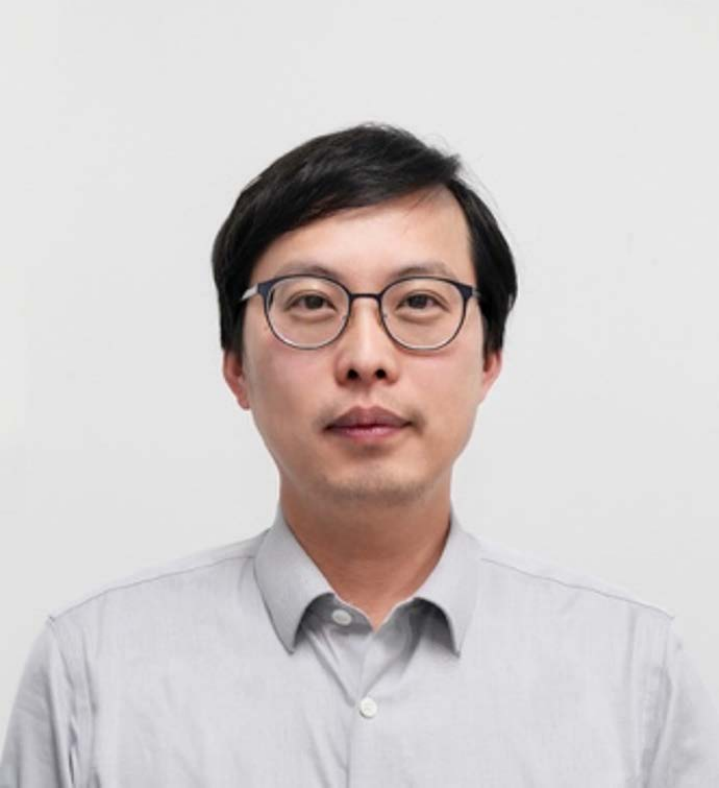}}]{Zhan Ma} (SM'19) is a Professor in Electronic Science and Engineering School, Nanjing University, Nanjing, Jiangsu, 210093, China. He received the B.S. and M.S. degrees from the Huazhong University of Science and Technology, Wuhan, China, in 2004 and 2006, respectively, and the Ph.D. degree from the New York University, New York, in 2011. From 2011 to 2014, he has been with Samsung Research America, Dallas, TX, and  Futurewei Technologies, Inc., Santa Clara, CA, respectively. His research focuses on learning-based video communication and computational imaging. He is a co-recipient of the 2019 IEEE Broadcast Technology Society Best Paper Award, the 2020 IEEE MMSP Image Compression Grand Challenge Best Performing Solution, the 2023 IEEE WACV Best Algorithm Paper Award, and the 2023 IEEE Circuits and Systems Society Outstanding Young Author Award.
\end{IEEEbiography}

\begin{IEEEbiography}[{\includegraphics[width=1in,height=1.25in,clip,keepaspectratio]{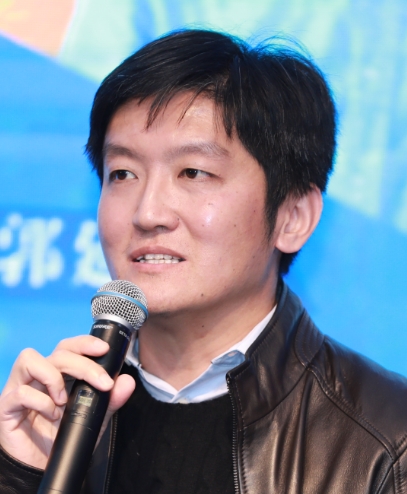}}]{Yanwen Guo} received the PhD degree in applied mathematics from the State Key Lab of CAD\&CG, Zhejiang University in 2006. He is a professor with the National Key Lab for Novel Software Technology, Nanjing University. He was a visiting scholar with the Department of Electrical and Computer Engineering, University of Illinois at Urbana-Champaign, from 2013 to 2015. His research interests include image and video processing, vision, and computer graphics.
\end{IEEEbiography}

\begin{IEEEbiography}[{\includegraphics[width=1in,height=1.25in,clip,keepaspectratio]{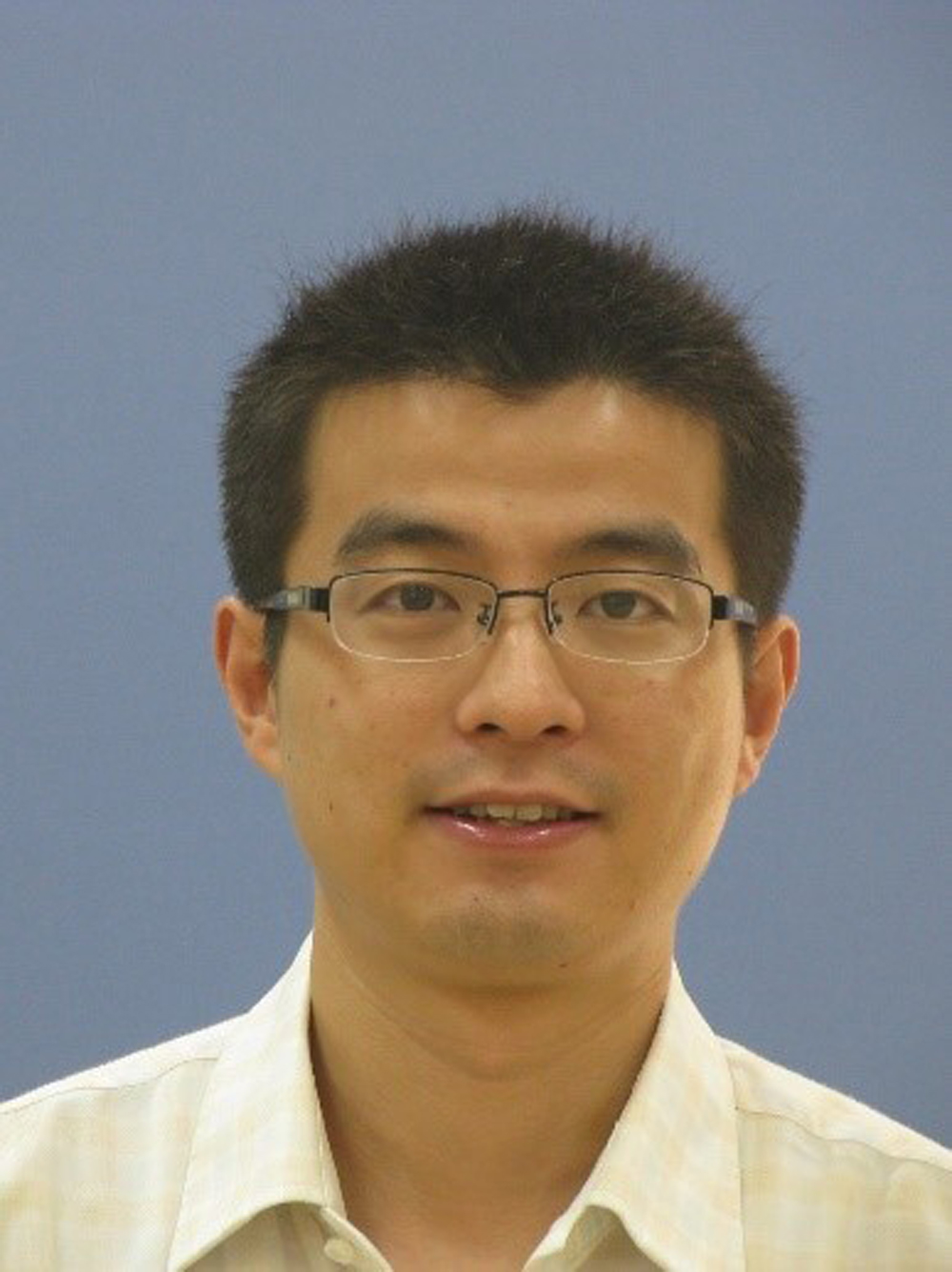}}]{Xun Cao} received the B.S. degree from Nanjing University, Nanjing, China, in 2006, and the Ph.D. degree from the Department of Automation, Tsinghua University, Beijing, China, in 2012. He held visiting positions with Philips Research, Aachen, Germany, in 2008 and Microsoft Research Asia, Beijing, from 2009 to 2010. He was a Visiting Scholar with the University of Texas at Austin, Austin, TX, USA, from 2010 to 2011. He is a Professor at the School of Electronic Science and Engineering, Nanjing University. His current research interests include computational photography and image-based modeling and rendering.
\end{IEEEbiography}




\end{document}